\documentclass[final,3p,times,onecolumn]{elsarticle}

\usepackage{graphicx}
\usepackage{xcolor}
\usepackage{listings}
\usepackage{caption}
\usepackage{tabularx}		%% For simple table stretching
\usepackage{threeparttable}
\begin{document}

\begin{frontmatter}
\title{Towards a Rigorous Evaluation of Explainability  \\ for Multivariate Time Series}

\author[Aalto,KTH]{Rohit Saluja\corref{cor1}}
\cortext[cor1]{First and Corresponding author}
%\ead{rohit.saluja@aalto.fi}
\author[bournemouth,Aalto]{Avleen Malhi}
%\ead{avleen.malhi@aalto.fi}
\author[Aalto,Umea]{Samanta Knapič}
%\ead{samanta.knapic@aalto.fi}
\author[Aalto,Umea]{Kary Fr\"{a}mling}
%\ead{kary.framling@cs.umu.se}
\author[KTH]{Cicek Cavdar}
%\ead{cavdar@kth.se}

\address{\{rsaluja@kth.se, avleen.malhi@aalto.fi, samanta.knapic@aalto.fi, kary.framling@cs.umu.se, cavdar@kth.se\} }
\address[Aalto]{Department of Computer Science, Aalto University, Konemiehentie 2, 02150 Espoo, Finland}

\address[Umea]{Department of Computing Science, Ume\r{a} University, Mit-huset, 90187 Ume\r{a}, Sweden}
 
\address[bournemouth]{Department of Computing and Informatics, Bournemouth, United Kingdom}
 
\address[KTH]{Department of Electrical Engineering and Computer Science, KTH Royal Institute of Technology, Stockholm, Sweden}

\begin{abstract}
%Machine learning-based systems are rapidly becoming popular due to the realization that machines are more efficient and effective than humans at performing certain tasks. Despite their popularity, they are also very literal and making predictions using machine learning algorithms only, can solve most real-world problems just partially. This has led to a huge research on machine learning explainability in order to ensure that machine learning models are reliable, fair, and can be held liable for their decision-making process. 
Machine learning-based systems are rapidly gaining popularity and in-line with that there has been a huge research surge in the field of explainability to ensure that machine learning models are reliable, fair, and can be held liable for their decision-making process. Explainable Artificial Intelligence (XAI) methods are typically deployed to debug black-box machine learning models but in comparison to tabular, text, and image data, explainability in time series is still relatively unexplored. The aim of this study was to achieve and evaluate model agnostic explainability in a time series forecasting problem. This work focused on proving a solution for a digital consultancy company aiming to find a data-driven approach in order to understand the effect of their sales related activities on the sales deals closed. The solution involved framing the problem as a time series forecasting problem to predict the sales deals and the explainability was achieved using two novel model agnostic explainability techniques, Local explainable model- agnostic explanations (LIME) and Shapley additive explanations (SHAP) which were evaluated using human evaluation of explainability. The results clearly indicate that the explanations produced by LIME and SHAP greatly helped lay humans in understanding the predictions made by the machine learning model. The presented work can easily be extended to any time series forecasting or classification scenario for achieving explainability and evaluating the produced explanations.
\end{abstract}

\end{frontmatter}

\section{Introduction}
Machine learning is rapidly entering into almost every industry as it has been realized that machines are faster, more efficient, and more effective than humans at performing specific tasks \cite{doshi2017towards}. Industries, where machine learning has gained a lot of popularity, include the healthcare, financial, retail, and transportation industries. However, building predictive machine learning models that capture complex relationships of the data, using them to make predictions, and evaluating the models using an accuracy metric only solves a part of the problem. Doshi-Velez and Kim \cite{doshi2017towards} in their research work state that building a machine learning model that performs well at making decisions only offers an incomplete description of most real-world tasks. It is also important to understand the why behind the decision as opposed to just finding out the decision being made by a machine learning model. Another problem highlighted by Ribeiro et al. \cite{ribeiro2016should} is that humans as a curiosity-driven species have a desire to understand the reasoning behind the decisions made by a machine learning model and will therefore not use a machine learning model that they can not fully understand or trust. 

Due to these reasons, recent years have witnessed a lot of criticism towards machine learning. This has pushed researchers to work on understanding the decisions made by the machine learning models. One way that can help understand their decisions is by using explainability techniques. According to Miller \cite{miller2017explanation}, explainability is defined as “the degree to which a human can understand the cause of a decision”. Explainability in machine learning can be achieved in different ways, either by building machine learning models that are inherently explainable due to their simple structure, such as models built with linear regression, logistic regression, and short decision trees, or by interpreting a model after it has been trained with so called model agnostic or post-hoc explanation tools \cite{molnar2020interpretable}. Another distinction can be made between model-specific tools that only work with specific intrinsically explainable models and the model agnostic tools that are used to interpret models post-training (post-hoc) and can be used to interpret any machine learning models, usually complex in nature and also referred to as black-box models \cite{molnar2020interpretable}. Explainability tools can produce two kinds of model interpretations, either local or global. Global explainability helps in understanding the overall effect of input values on the output of the machine learning model, whereas Local explainability helps in understanding the effect of input values on the output produced by the machine learning model for a particular instance in the data set. A good human-understandable output produced after interpreting the model is called an explanation \cite{molnar2020interpretable}.

\begin{figure*}[!h]
\centering
\begin{minipage}{.70\textwidth}
  \centering
  \includegraphics[width=0.90\linewidth]{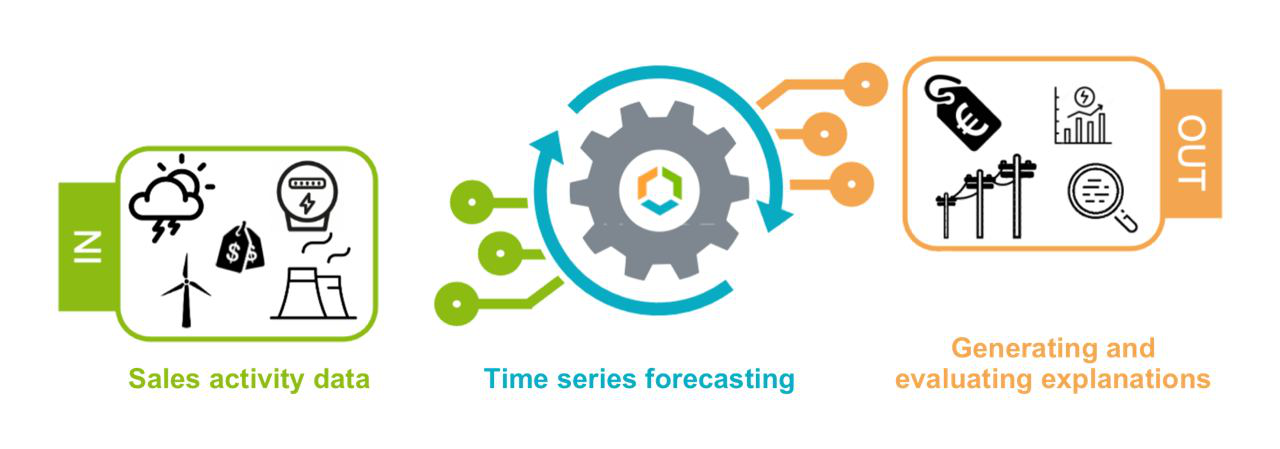}
  \captionof{figure}{Role of explainability in time series forecasting}
  \label{fig:intro}
\end{minipage}%
\end{figure*}

The different types of data that are fed into machine learning algorithms can be categorized into five broad categories - numerical data, categorical data, text data, image data, and time series data. Time series contains data points measured over a period of time and can be differentiated from the other data types based on the dependence of the dimension of time \cite {esling2012time}. Time series holds a lot of importance because it is the most dominant type of data being generated in the fields of business, economics, science, and engineering as most of the measurements are performed over time. Due to the presence of time series data in every major industry, there is an increasing need for understanding the decisions made by models trained on time series data. However, explainability with time series data can be challenging due to the complex non-linear temporal dependencies in the data. The elements of time series data are connected to each other with the time dimension and the sequence of the data in a time series data becomes extremely important. While working with time series data, just identifying the features leading to the decision of a machine learning model is not enough. It is also extremely important to identify how the features from different points in time affect the decision of the machine learning model \cite{hidasi2011shifttree}.
Simple models like linear regression and short decision trees can be used to achieve explainability but they might not be sufficient to capture complex relationships of the underlying data, whereas complex black-box machine learning models can be used to achieve good accuracy on the data but might not be explainable. This leads to a tradeoff between accuracy and explainability \cite{molnar2020interpretable}. Using tools that are specifically built for achieving explainability of certain models can hinder the flexibility of trying out different models for a good performance. The role of explainability in time series forecasting is shown in Figure \ref{fig:intro}. To address the aforementioned concerns, it is important to think in terms of model agnostic methods for the time series. 

Based on that the first research question being addressed in the present work is: How to achieve model agnostic explainability in a time series forecasting problem?
Further the ultimate goal of explainability is to aid humans in understanding the decision-making process of machine learning models. As highlighted by Doshi-Velez and Kim \cite{doshi2017towards}, different kinds of explanations produced after interpreting machine learning models may not be equally explainable. Moreover, their research work claims that, with the growing research in the field of explainability in machine learning, there is also a pressing need to be able to quantify the quality of explanations produced after interpreting the machine learning models. Due to these reasons, the evaluation of explainability is extremely important and the research work in this field is gaining a lot of momentum. However, the evaluation of explainability for machine learning models built with time series data is still unexplored, especially in the context of a time series forecasting problem.
In connection to this our second research question being addressed in the present work is: How to evaluate evaluate the explanations produced by interpreting the machine learning time series forecasting model? 

%A time series that consists of a single value sequentially measured over a period of time is called a univariate time series whereas a time series that consists of two of more values sequentially measured over a period of time is called a multivariate time series [6, p. 16]. Time series forecasting is the process of understanding the past and using it to predict the future. During the process of time series forecasting, the historical observations of the time series are used to build models capable of predicting the future values of the series. 

We  addressed the research questions  in a study conducted with a company called Futurice in Helsinki, Finland under their Exponential AI team  driving the objective of making Futurice a data-driven organization. Futurice being a consultancy company is highly dependent on the sales of its consultancy services for running the business. However, the company has zero visibility into the aspects that drive its sales and because of that, the current system of understanding, whether they are getting enough sales activity in the company, is completely hunch based. In an attempt to become a more data-driven organization, Futurice wants to understand the effect of various sales-related activities conducted in the company on the sales deals closed by the company. Naturally, these various sales activities have sequentially occurred across time and have a time dimension attached to them. This makes the type of the data time series. Using a data-driven strategy to understand the effect of the sales activities on the sales of the company can be used to inform the strategy of the sales teams at the company. A data-driven sales strategy can also help weed out human bias from the decision-making process. 
The purpose of the present work was therefore to build a framework for achieving model agnostic explainability in a multivariate time series forecasting problem and to evaluate the explanations produced by interpreting the machine learning model for time series forecasting. The goal was to achieve explainability on a predictive model with multivariate time series data in order to understand the decision-making process of the underlying model and draw useful actionable insights. Our goal can be divided into four sub-goals:
\begin{enumerate}
\item Preparing the relevant time series data and formulating the forecasting problem.
\item Building time series forecasting models with different lag variables and selecting the best model based on an evaluation metric.
\item Interpreting the time series forecasting model to produce human understandable explanations.
\item Evaluating the explanations produced.
\end{enumerate}

\section{Background}

%Any work done in the field of explainability has a huge ethical and societal impact attached to it. In fact, one of the main reasons why research in the field of explainability is growing at a rapid pace is because of the realization that just building highly accurate machine learning models is not good enough. It is equally important to ensure that these models are reliable and fair. Machine learning algorithms are extremely powerful, but they are also very literal. For example, a machine learning model tasked to find the cheapest possible cure for cancer would probably find ways to end lives rather than saving them. Another example with respect to ethics as mentioned in [1, p. 3], talks about the need for an unbiased loan approval classifier. A loan approval classifier should not discriminate against people on the basis of race, ethnicity, or gender. Due to ethical and societal concerns, the European Union passed the right to explanation act [11] which states that all algorithmic systems should be able to provide explanations. 

%Time series data is extremely common and is largely present in the field of economics, business, science, and technology. The work done during this thesis is shedding light on explainability in time series and has the potential to have a huge impact in terms of addressing ethical and societal issues. 

\subsection{Time series data forecasting}
\begin{figure*}[!h]
\centering
\begin{minipage}{.40\textwidth}
  \centering
  \includegraphics[width=0.50\linewidth]{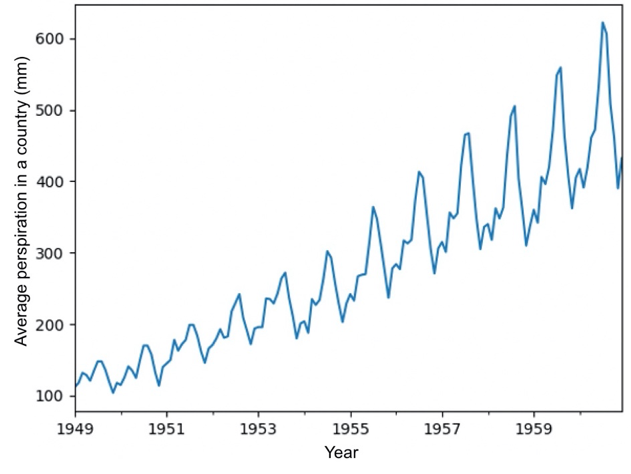}
  \captionof{figure}{Average perspiration in a country against year\cite{brownlee2017introduction}.}
  \label{fig:uni}
\end{minipage}%
\centering
\begin{minipage}{.60\textwidth}
  \centering
  \includegraphics[width=0.70\linewidth]{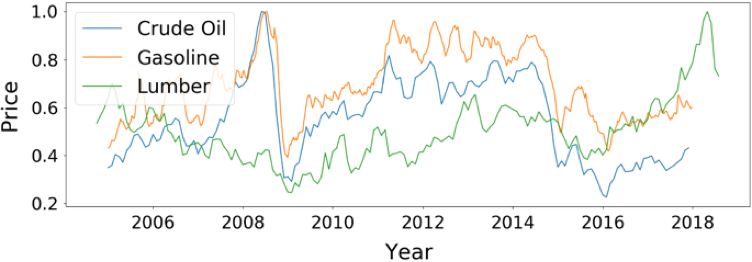}
  \captionof{figure}{Price of Crude Oil, Gasoline and Lumber against a year\cite{shih2019temporal}.}
  \label{fig:multivariate}
\end{minipage}%
\end{figure*}

The time series data is formed by the data points that are sequentially measured over a period of time form. If the data is generated at regular intervals of time it forms a regular time series whereas the data generated at an irregular interval of time forms an irregular time series. %The most common examples of time series data are stock prices recorded at the end of every day, sales of a company recorded at the end of every month, revenue generated by a company every month, hourly average temperature readings of a particular location, reading generated by a heart rate monitor every second, the data generated by various sensors in mobile phones, data generated by autonomous vehicles about its environment, reading recorded by an IoT sensor during an experiment, electrocardiogram (ECG) data generated in the field of medicine, etc. 
A time series that consists of a single value sequentially measured over a period of time is called a univariate time series (Figure \ref{fig:uni}, adapted from \cite{brownlee2017introduction}), whereas a time series that consists of two or more values sequentially measured over a period of time is called a multivariate time series (Figure \ref{fig:multivariate}, adapted from \cite{shih2019temporal}).
The historical observations of the time series are used to build models capable of predicting the future values of the series \cite{brownlee2017introduction} during the process of time series forecasting. Time series forecasting is the process of understanding the past and using it to predict the future. In the time series nomenclature the current state of time is defined with subscript \textbf{t} and the observation at the current state of time is described as \textbf{obs (t)}. Times before the current time are considered negative relative to the current time and are called lag times and the observations at those times are called lagged observations. Similarly, the time in the future is considered positive with respect to the current time t, and the observations at those times are called lead time:
\begin{itemize}
%\item The current time which is used as the time of reference is represented by \textbf{t}.
\item The observations at \textbf{t-1} and \textbf{t-2} would be \textbf{obs(t-1)} and \textbf{obs(t-2)} respectively.
\item The previous or lagging time steps are represented by \textbf{t-n}, where n is a particular time instance (e.g. t-1, t-2, t-3, and henceforth) 
\item The future or leading time steps are represented by \textbf{t+n}, where n is a particular time step (e.g. t+1, t+2, t+3, and henceforth).
\item The observations at time instance \textbf{t+1} and \textbf{t+2} are represented by \textbf{obs(t+1)} and \textbf{obs(t+2)} [6].
\end{itemize}

Because time series forecasting can be framed as a supervised learning problem, this enables the use of various linear and nonlinear machine learning algorithms to predict values at future time steps. 
%This is achieved using the values from the previous time steps in a time series to predict the values in the future time steps [6, p. 14]. 
Our time series data set was trained using Support Vector Regression (SVR), a supervised learning method, commonly used for machine learning problems of classification type. 
%SVM projects the training data into a higher feature dimensional space using a kernel function and finds an optimal hyperplane to create a decision boundary with the maximum possible margin. %Projecting the training data into a higher-dimensional space makes it easier to find an optimal hyperplane for classification. 
%Different kernel functions available in SVM include Linear, Polynomial, and Radial Basis Function (RBF) kernel. To find the optimal decision boundary SVM uses support vectors, the data points that lie on the maximized margin \cite{gunn1998support}.
%The solid black line in the center is the best hyperplane that separates the classes C1 and C2. The best hyperplane is found by optimizing for the maximum margin distance (best margin) with the help of the support vectors that are represented by dotted lines in the figure. 
%SVM was initially developed for classification problems but was later extended to suit regression problems. 
SVR operates with the same logic as the Support Vector Machine (SVM) but rather than finding a decision boundary with maximum possible margin SVR focuses on finding an approximate function to minimize the error of the loss function. SVR tries to find a decision boundary depending on the defined loss function that ignores errors that are located within a particular distance of the true value. 
%The distance is denoted with a variable $\epsilon$. 
Hence, SVR does not care about the whole training data and depends on a subset of training data to make predictions. 
%The $\epsilon$ variable dictates the tolerance for error in SVR. 
The concept of kernels also holds true for SVR and nonlinear data is mapped into a data of higher dimensional space to make it linear \cite{drucker1997support}.  
%In Figure \ref{fig:svm2} \cite{drucker1997support}, the orange line represents the function built by SVR and the green dotted lines form the $\epsilon$ tube (margin). The points outside the $\epsilon$ tube, highlighted with black circles are the support vectors in SVR. The best fit hyperplane will try to have the maximum number of data points from the training data inside the $\epsilon$ tube. 

\begin{figure*}[!h]
\centering
\begin{minipage}{.50\textwidth}
  \centering
  \includegraphics[width=0.70\linewidth]{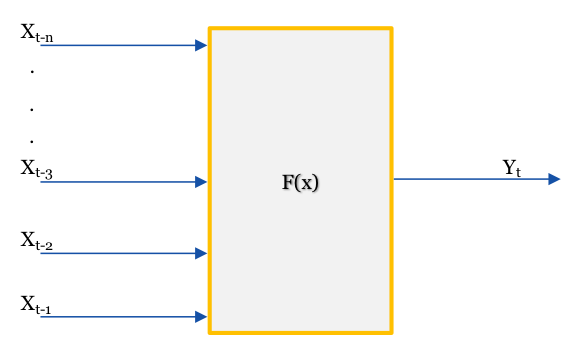}
  \captionof{figure}{Time series forecasting as a supervised learning problem.}
  \label{fig:supervised}
\end{minipage}%
\end{figure*}

In the supervised machine learning an algorithm tries to learn a mathematical function that maps the input variable (input feature) X to an output (target) variable Y as accurately as possible. This learned mapping function is then used to predict the output for new input variables. In one step forecasting, the values from the previous time steps are used to predict the value at the next time step. In multi-step forecasting the values from the previous time steps are used to predict values at two or more future time steps.
Figure \ref{fig:supervised} depicts the procedure of approaching a unit step time series forecasting as a supervised machine learning problem. The set of input features formed using the lag values of a univariate time series are denoted by Xt-1, Xt-2, Xt-3...Xt-n where n is denoted by the final lag value or the size of the lag and is less than the total of data points in the data set. The machine learning model is denoted by F(x) and the output, Yt is the predicted value at a future time step. Similarly, the concept can be extended to multivariate time series and lag values of different variables can be used to predict the values at the future time steps.

%\begin{figure*}[!h]
%\centering
%\begin{minipage}{.39\textwidth}
%  \centering
%  \includegraphics[width=0.59\linewidth]{figures/SVM.png}
%  \captionof{figure}{Intuition behind SVM in a binary classification case}
%  \label{fig:svm}
%\end{minipage}%
%\begin{minipage}{.60\textwi%  \centering
%  \includegraphics[width=0.70\linewidth]{figures/SVM2.png}
%  \captionof{figure}{Intuition behind a two-dimensional SVR [17].}
%  \label{fig:svm2}
%\end{minipage}%
%\end{figure*}

%The various parameters that can be tuned in SVR are [16, p. 19-23]:
%\begin{itemize}
%\item 	Kernel - Linear, Polynomial or Radial Basis Function Kernel 
%\item C – The penalty factor to the error of the loss function (regularization parameter)
%A higher C penalizes the error more and could lead to low generalization (high overfitting) of the function built by SVR. A lower C penalizes the error loss. 
%\item Size of the $\epsilon$ tube (margin)
%\item Gamma, if the Kernel is Radial basis function (RBF) 
%\end{itemize}

\subsection{Concepts of explainability}
%Recently with the machine learning and data science entering every domain of life, a lot of research in the field of explainability in machine learning has been happening in order to enable humans to thoroughly understand the reasons behind the classification or regression outputs of these machine learning models.
Definition of explainability offered by Kim et al. \cite{kim2016examples} in their research work is that “explainability is the degree to which a human can consistently predict a model’s output”. Explainability helps humans to better understand the following elements of the decision making of a machine learning model \cite{doshi2017towards}:
\begin{itemize}
\item It helps to understand the fairness of a machine learning model. 
%For example, the explainability of a model that approves or rejects a loan application at a bank would help the humans understand if the model is or is not discriminating based on gender or race. 
\item It ensures that the privacy is not violated in any way and that the sensitive information in the data is protected. 
\item It helps in understanding the reliability and robustness of a machine learning algorithm. %In a robust machine learning model, the prediction does not drastically change if a small amount of noise is added to the input data.
\item It generates trust by enabling humans to understand the reasons behind a machine learning model's decisions. 
\end{itemize}

\begin{figure*}[!h]
\centering
\begin{minipage}{.50\textwidth}
  \centering
  \includegraphics[width=0.70\linewidth]{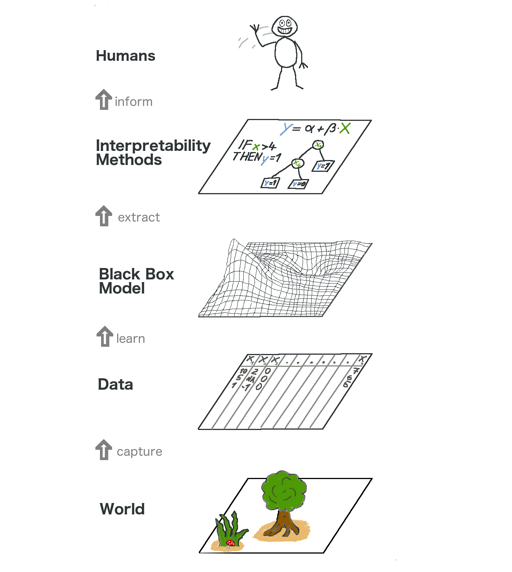}
  \captionof{figure}{Model agnostic explainability [4].}
  \label{fig:agnostic}
\end{minipage}%
\end{figure*}

In the present work we focused  on the Model Agnostic explainability. Model Agnostic tools do not acquire any internal information about the inner working of the model like its weights or any other structural information \cite{molnar2020interpretable}. While the intrinsically explainable models are not capable of picking up complex relationships between the input features and the target variable, the model agnostic explainability tools help overcome the trade off between models' performance and explainability. Another great advantage of using model agnostic explainability methods is that explainability of different machine learning models used for the same task can be compared \cite{ribeiro2016model}. The overall picture of model agnostic explainability is presented in Figure \ref{fig:agnostic} and was adapted from \cite{molnar2020interpretable}. 

Molnar \cite{molnar2020interpretable} in his work discusses various model agnostic tools which include partial dependence plot, individual condition expectation, accumulated local effects, feature interaction, permutation feature importance, and surrogate methods like LIME and SHAP. The surrogate model is an intrinsically explainable model and the goal of a good surrogate model is to approximate the black box prediction model as accurately as possible. Since the surrogate model only requires the prediction function of the black-box model and the data without having any information regarding the working of the black-box model, it is model agnostic. The biggest advantage of model agnostic techniques like LIME and SHAP is that they are feature attribution methods, meaning that they present the explanation of a complex (black box) machine learning model in terms of the contribution of features leading to prediction. 
One of the biggest advantages of SHAP is also its solid theoretical foundations in game theory (Shapley values) which ensures that prediction is distributed fairly among features. Additionally the explanations of feature attribution methods are extremely human friendly and can make sense to lay humans. 

%Molnar in [4, p. 201] further summarizes the steps used to build a global surrogate model and approach explainability using surrogate models.
%\begin{itemize}
%\item A dataset is selected. This dataset could be the same as the one used to train the black-box model or a subset of the original dataset like the validation or test set. 
%\item The predictions for the selected dataset are obtained using the black-box model. 
%\item An interictally explainable model like linear regression or decision tree is trained on the selected dataset and the predictions from the black-box model. This results in the surrogate model. 
%\item The surrogate model is then interpreted to approximately explain the effect of various input features on the target prediction of the black-box model. %\end{itemize}

%Since the surrogate model is trying to approximate the black-box model and not make predictions using the original data, the prediction quality of the black box model influences the quality of interpretations obtained from the surrogate model. If the black-box model is performing poorly then the conclusions drawn about the black box model in form of interpretations using the surrogate model would also be poor [4, p. 202].  

\subsubsection{Local explainable model agnostic explanations (LIME)}
Ribeiro et al. \cite{ribeiro2016should} in their research work proposed a local surrogate model called LIME to explain the prediction on a single instance of data by a black box machine learning model. In comparison to global surrogate models which focus on explaining the holistic effect of features on the predictions made by the black-box model, a local surrogate model like LIME focuses on explaining individual predictions. LIME follows the following general approach to explain individual predictions \cite{molnar2020interpretable}:
\begin{itemize}
\item After the instance whose prediction needs to be explained is provided to LIME, it permutes the data set to create new sample data. 
\item Corresponding predictions for the new sample data are obtained using the black-box model. 
\item The new sample data is weighted with respect to its proximity to the instance whose prediction needs to be explained. 
\item A weighted explainable model is trained on the sample data and the corresponding predictions.
\item The trained explainable model is interpreted in order to explain the individual prediction. 
\end{itemize}

%The local surrogate model should be good at approximating the black-box model locally but does not necessarily have to be good at approximating the model behavior globally. 
%Figure \ref{fig:LIME} from the paper [2, p. 1138] presents a toy example to highlight the intuition behind LIME. The blue/pink color represents the complex decision boundary of the black box prediction function. The dark red cross is the instance that needs to be explained. LIME creates new sample data and obtains its predictions using the black box model which are then weighted with respect to their proximity to the instance that needs to be explained. These weights are represented by the size in the image. The dashed line represents the obtained surrogate model like linear regression that can be interpreted to explain the prediction [2, pp. 1137-1138].

%\begin{figure*}[!h]
%\centering
%\begin{minipage}{.50\textwidth}
%  \centering
%  \includegraphics[width=0.70\linewidth]{figures/lime-method.png}
%  \captionof{figure}{The intuition behind LIME highlighted for a black box model for binary classification [2]}
%  \label{fig:LIME}
%\end{minipage}%
%\end{figure*}

Mathematically LIME’s explanation of a local and individual prediction can be expressed in the form of Equation \ref{eqn:Lime} \cite{molnar2020interpretable}: 
\begin{equation}
explanation(x) = argmin_{g \in G}\mathcal{L}(f,g, \pi_x) +  \, \Omega(g)
\label{eqn:Lime}
\end{equation}  
The explanation of $x$ result of maximisation of fidelity term $\mathcal{L}(f,g, \pi_x)$ with complexity of $\Omega(g)$. $f$ represents a black-box model which is explained by explainer represented by $g$. The local surrogate model attempts to locally fit the data in the proximity of the instance whose prediction is to be explained. Fitting the local model requires enough data around the vicinity of instance being explained, which is done by sampling the data from its neighbourhood \cite{molnar2020interpretable}.

%The technical details behind how the size of the neighborhood is decided can be found in [4, pp. 209-210]. 

\subsubsection{Shapley Additive explanations (SHAP)}
SHAP (SHapley Additive exPlanations) [14] is a surrogate model approach to interpret black box models. SHAP is a method based on a concept from cooperative game theory called Shapley values [20]. SHAP offers local explanations with the shapley value-based method to explain the cause of individual predictions and also offers global explainability based on the addition of Shapley values from individual predictions.

SHAP \cite{lundberg2017unified} aims to explain individual predictions by employing the  game-theoretic Shapley value \cite{shapely1953value}. This approach uses the concept of coalitions in order to compute (as shown in equation \ref{eqn:Shap1}) the Shapley value of features for the prediction of instance ($x$) by the black-box model $(f)$. The Shapley value is the average marginal contribution $(\phi_j^m)$  of feature $(j)$ in all possible coalitions.  The marginal contribution is calculated as in equation \ref{eqn:Shap2} where $\hat{f}(x_{+j}^m)$ and  $\hat{f}(x_{-j}^m)$ are prediction of black-box $f$ with and  without the $j^{th}$ feature of instance $x$ from the sample. 
 \begin{equation}
 \phi_j(x)= \frac{1}{M} \sum_{m=1}^{M} \phi_j^m
 \label{eqn:Shap1}
\end{equation}

\begin{equation}
 \phi_j^m= \hat{f}(x_{+j}^m) - \, \hat{f}(x_{-j}^m) 
 \label{eqn:Shap2}
\end{equation}

%+++
%Mathematically, the shapley values estimation formula as defined by Lundberg and Lee in [14] and reiterated by Molnar in his book [4, p. 226] is as follows (Equation 2 2 [4, p. 226] [4, p. 226]).

\subsection{Explanations evaluation}
Until now, there is no clear consensus in the scientific community on how to evaluate explainability \cite{molnar2020interpretable}. However, initial research by Doshi-Velez and Kim \cite{doshi2017towards} suggests evaluation in the following ways:

\begin{figure*}[!h]
\centering
\begin{minipage}{.60\textwidth}
  \centering
  \includegraphics[width=0.80\linewidth]{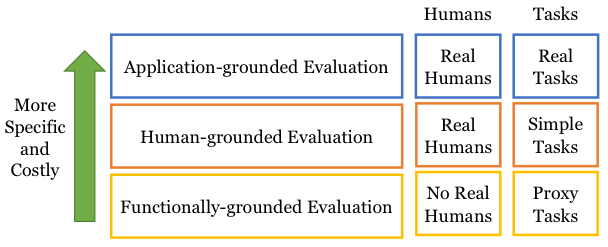}
  \captionof{figure}{Summary of the various evaluation methods of explainability along with their attributes.}
  \label{fig:eval}
\end{minipage}%
\end{figure*}

\begin{enumerate}
\item In the application level (grounded) evaluation, the explanations generated after explainability are tested by the domain experts/end user. %The quality of the evaluations is evaluated with respect to the end task. 
%For example, explanations for a machine learning model tasked with performing diagnosis of a particular disease would be evaluated by doctors performing the diagnosis. 
If the domain expert is able to understand and explain the decision made by the model, the interpretation can be considered as good. This evaluation technique is considered the best because it directly evaluates the explanations with respect to the final objective/task of the machine learning system. However, conducting application-level evaluation is very costly in terms of the time and effort involved and thus difficult to conduct. 
\item In human level (grounded) evaluation, the explanations are presented to ordinary people for evaluation. During this evaluation people are subjected to simple experimental tasks. Human-level evaluation is a metric that depends on the quality of the generated explanation and is independent of the prediction accuracy of the underlying machine learning model. This type of the evaluation is an attractive choice because it reduces the cost of experimentation and is also simpler to conduct as compared to application-level evaluation. 
%Some examples of possible experiments include:
%\begin{itemize}
%\item Forward simulation experiment where humans are presented with an input to the ML model and its explanation and then asked to correctly simulate the output of the ML model. 
%\item Counterfactual simulation experiment where humans are presented with an input to the ML model along with its output and explanation and asked about possible changes that can be made to the model’s input to change its prediction to a desired output. 
%\end{itemize}
\item Functionally-grounded evaluation does not require humans and is well suited for models that have been already evaluated by humans. It is cheaper than human-level evaluation in terms of time, effort, and cost. Function level evaluation involves a proxy task based on some formal definition of explainability to evaluate the quality of explanations which can be challenging to determine. 
\end{enumerate}

\section {Literature review}
\subsection{Interpretibility approaches in literature}
 Mokhtari et al. \cite{mokhtari2019interpreting} have approached the explainability of financial time series using SHAP. The problem tackled in this paper was a classification problem and both binary class and multi-class classification models were built using SVM, XGBoost, Random Forest, k Nearest Neighbors (kNN), and LSTM. The predictions made were interpreted using SHAP to understand the most important features for the prediction and to understand the contribution of the new data set in comparison to the old data set for the prediction task. kNN and SVM had the best performance on their data set and SHAP was useful in drawing crucial insights about the prediction.  
The research work done in \cite{mokhtari2019interpreting} only focuses on the global explanations and only experiments with SHAP as an explanation technique. The biggest limitation is that no methods were applied to evaluate the quality of the explanations produced after interpreting the model.

García et al. \cite{garcia2020shapley} in their research used multivariate time series data about various atmosphere related factors like wind speed, solar radiation, temperature, humidity, etc. to predict the NO2 concentrations in the city of Madrid using LSTM. The predictions were then interpreted using SHAP to understand the overall effect of features on particular prediction instances. SHAP was also used to obtain the feature importance to understand the overall impact of each feature on the prediction of NO2 concentrations. 
The research work done in \cite{garcia2020shapley} produces both local and global explanations for its time series forecasting model using SHAP. However, a  limitation of the work is that the quality of the explanations produced (local and global) was not evaluated.

%Malhi et al. \cite{malhi2020explainable} leveraged LIME explanations 
%to identify if autonomous agents accompanied with these explanations could help humans identify bias in the decision-making process of the autonomous agent. They concluded that these explanations increased the trust of the humans in the model and helped them in identifying biases posed by the autonomous agent. 

Madhikermi et al. \cite{madhikermi2019explainable} in  their research work elaborate how LIME could be used for explaining the heat recycler’s fault detection in Air Handling Unit (AHU). They used Support Vector Machines (SVM) and neural networks to build their classifier for the fault detection in AHU and then used LIME to explain the decision-making process of the underlying models. The XAI algorithm LIME was used to explain the prediction on 6 data instances. Their research work concluded that the LIME explanations add huge value in terms of increasing trust in the underlying SVM and neural networks classifier.

Schlegel et al. \cite{schlegel2019towards} in their research work applied various XAI techniques to time series. This included Saliency maps, LRP, DeepLIFT, LIME and SHAP. During their research they concluded that the XAI techniques that were originally proposed for text and image data do work with time series data as well. Based on their evaluation they concluded that SHAP was more robust for time series than LIME. These results are based on a novel verification method they seemed to have developed. This method of evaluating explanations derived from models built on time series seems to be unverified. The primary criteria for evaluating any kind of XAI explanation is by gauging if it improves the understanding of the decision-making process of the underlying model for humans. Another good factor to consider is if it leads to an increased trust in the model. Their verification based evaluation process for the explanations is not supported by any kind of human involvement which seems to a major drawback of the work.

\subsection{Evaluation of explainability for machine learning models}
The evaluation of explanations produced after interpreting the machine learning models built with time series data is currently still unexplored, especially in the context of a time series forecasting problem. Due to this reason, there is no direct point of reference for designing the human evaluation experiments. However, we thoroughly explored other research work related to the human evaluation of explanations.
For example, Yu et al. \cite{yu2020human} focused on the evaluation of the explanations by interpreting the music generated by an artificial intelligence (AI) system. Furher Nguyen \cite{nguyen2018comparing} conducted a forward prediction (simulation) task for human evaluation and evaluated different explanations produced by various explainability techniques in a text classification problem. Nguyen during her research worked with a binary classification case where the model predicts the sentiment of the movie review, positive or negative based on the text of the review. During the forward prediction task, Nguyen presented ordinary humans with the input to the model that is the text of the review and the explanation of the model. Based on the input and explanation, the humans were asked to predict the output of the model, negative or positive sentiment. 

Hase and Bansal \cite{hase2020evaluating} in their research work focus on the evaluation of explanations produced using various explainability techniques on a sentiment classification problem. They worked on the movie review text data set with positive or negative output labels and tried to improve the design of the forward prediction(simulation). Although using the forward prediction task in a binary classification setting is rational, it was not selected for the present study because using it in a regression (forecasting) setting seemed unreasonable, as well as asking ordinary humans to predict a numerical output on the basis of input feature and explanation because there are infinite numbers in the number system.

Malhi et al. \cite{malhi2020explainable} conducted a human evaluation of explainability for an machine learning model built for approving or rejecting a bank loan application using various attributes of the loan applicants. The researchers here employed a verification task by presenting the humans with an input, an output (approve or reject), and an explanation. Based on the input, output, and explanation, the humans were asked to agree or disagree with the output of the model. It was realized that a variation of verification task could be used for human evaluation of the produced explanations in the case of a time series forecasting problem.

\section{Methods}
\begin{figure*}[!h]
\centering
\begin{minipage}{.80\textwidth}
  \centering
  \includegraphics[width=1.0\linewidth]{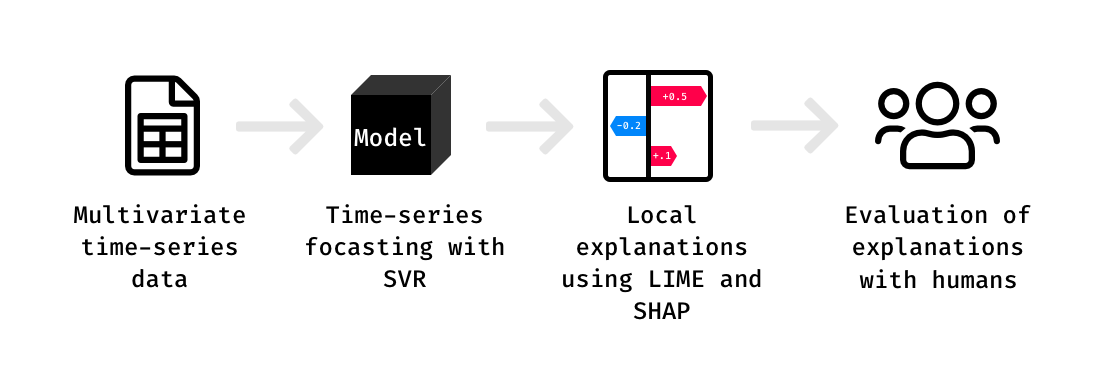}
  \captionof{figure}{Overview of the methodology used to address the research problems.}
  \label{fig:methodology}
\end{minipage}%
\end{figure*}

Due to the high complexity of machine learning and explainability methods involved our empirical research methodology can be divided into the five segments depicted in the Figure \ref{fig:methodology}. 
The first step involved understanding the sales processes of the company and the data streams that encapsulate the sales activity of the company. A qualitative analysis was conducted in form of interviews to achieve the aforementioned objectives. After the features were created, an exploratory data analysis was conducted to understand the features. After understanding the features, it was decided to approach the time series forecasting part of the project as a supervised learning problem using machine learning. When the data set was prepared for the forecasting it was fed into the support vector regression algorithm.  On top of the predictions made by the SVR model we implemented post-hoc explanation techniques Local explainable Model-agnostic Explanations (LIME) and SHapley Additive exPlanations (SHAP) in order to understand the influence of features on the predictions at a local level. The last task was the evaluation of the explanations produced after interpreting the SVR model using SHAP and LIME with the human-evaluation user study.

\subsection{Data collection}
In order to understand the sales processes of the company and the data streams that encapsulate the sales related activities of the company, the first step undertaken was conducting the interviews. The methodology of conducting interviews best aligned with the objective of obtaining the required domain knowledge to build useful features and prepare the list of sales activity related data streams.
%The company encourages employees to log data about various sales related activities in the company but due to self-organized culture of the company, the members of the sales teams do not necessarily log all the data about the sales process. Another important objective of these meetings was also to understand the sales activity related data that mandatorily gets logged by the sales teams throughout the company.  
The interviews were conducted with sales professionals, sales team leads, and the executives that are involved in the sales processes, with at least one year of sales experience in the company. The interviews were conducted in a semi structured format as it enables the interviewee to answer a particular set of questions but also enables them to share their opinions, thoughts and experiences freely \cite{dearnley2005reflection}. The interviews began with a fixed set of questions regarding the sales process, the type of tools used during the process, the type of data logged in the company’s system during the sales process, and the frequency of data logging. A total of 8 interviews were conducted and the names of the people have been anonymized due to privacy reasons. Table \ref{tab:inter} presents the statistics of the interview process conducted.

\begin{table}[!h]
\scriptsize
\centering
\caption{Statistics of the interviews to understand the sales process.}
\begin{tabular}{l|l|l}
\hline
Role at the Company                          & Time of employment at the company & Interview time   \\ \hline
Sales professional 1 in Helsinki office      & \textit{2 years}                  & \textit{30 mins} \\ \hline
Sales professional 2 in Helsinki office      & \textit{1 year}                   & \textit{30 mins} \\ \hline
Senior sales manager in Helsinki office      & \textit{3 years}                  & \textit{45 mins} \\ \hline
Sales team lead in Helsinki office           & \textit{5 years}                  & \textit{30 mins} \\ \hline
Sales professional 3 in Berlin office        & \textit{3 years}                  & \textit{45 mins} \\ \hline
Senior strategy executive in Helsinki office & \textit{6 years}                  & \textit{1 hour}  \\ \hline
Senior executive in Helsinki office          & \textit{19 years}                 & \textit{1 hour}  \\ \hline
\end{tabular}
\label{tab:inter}
\end{table}

\subsubsection{Data set preparation}
The next step involved procuring the data from the data streams and cleaning them to create usable features for machine learning. The understanding gained from the interviews was the basis for building the relevant data set for the machine learning forecasting model. Figure \ref{fig:data} illustrates the process of the method used for building the relevant sales activity data set holistically.
The process was completed using the following steps:

\begin{figure*}[!h]
\centering
\begin{minipage}{.70\textwidth}
  \centering
  \includegraphics[width=0.9\linewidth]{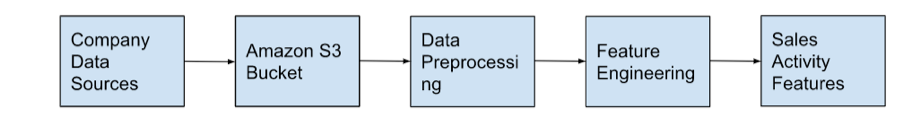}
  \captionof{figure}{Holistic view of the process of building the data sets.}
  \label{fig:data}
\end{minipage}%
\end{figure*}

\begin{enumerate}
\item Collecting the data from various data streams, for example the company’s data hosted on Amazon Web Services (AWS)’s Simple Storage Space (S3) bucket. 
\begin{itemize}
\item %The first step involved
The names of the people, involved in the sales at the company. 
%The company has a centralized human resources database that contains the names of the people who are currently employed by the company alongside their other details like email, phone number, pay scale, time of employment, and competence/role of the employee in the company. 
%This data was procured from the company’s data lake in Comma Separated Values (CSV) format and then the names of the employees with sales as a competence in the competence field of the dataset were extracted. 
%However, the problem was that this dataset only contained the names of the active employees of the sales teams and did not contain any information about the employees who have left the company. In order to make up for that and to complete the list of people with sales as a competence in the company, another database with information about the past employees of the company was manually procured from the human resources department. This database was made available in an excel spreadsheet format. It was then used to extract the names of the company’s past employees with sales competence. The final list of the employees with sales competence was then built. It contained the names of both the current and previous employees of the company. 
\item The data about the top fifty clients of the company in the last ten years, based on the total billing amount generated by the clients. 
%The billing records of the clients are stored in the financial database of the company. The data in the financial database of the company is manually updated on a monthly basis and pushed to the data lake of the company. 
%The client billing data procured from the data lake in the CSV format contained the following data fields: name of the client and billing amount.  
\item 
The data about booking internal and external meetings using Google calendar, involving the name of the organizer, the start date of the meeting, the title of the meeting, and the description text of the meeting.
%The data about meetings from Google’s calendar API is parsed and stored in the company’s data lake on a daily basis. The google calendar API offers a very exhaustive list of fields generated while creating a calendar invite. 
%The calendar data was obtained from the data lake in CSV format. 
\item The data about the hours spent working by the employees containing the  name of the employee, task completed, and the time taken to complete the task. 
%The company’s data lake pulls this data from the hour making software at the company on a daily basis. 
\item 
%The data of the CRM software was procured from the data lake in CSV format. 
The data of the Customer Relationship Management (CRM) software called Hubspot to record the journey of sales deals. Time of the deal and the status of the deal were extracted. 
\item The data about the expenses of the credit card used for various company-related expenses. The expenses data contains the name of the employee, transaction amount, and description of the transaction.
%The expenses are manually logged and stored by the employees in a centralized excel file at the end of every month. 
%The excel files are processed and made available on the company’s data lake. 

%was obtained from the data lake in the CSV format and 
\end{itemize}
\item Data preprocessing: The data was made error-free by removing empty or corrupted values. The formatting of the time data was fixed, and it was standardized to Eastern European time (EET). Date time columns were moved to the index location in time series for easy manipulation of the data. The irregular time series were converted into regular time series by aggregating the data to a monthly level. 
\item Feature engineering: After the features were created an exploratory data analysis was conducted. The domain knowledge collected during the interviews was used to create usable sales activity features for the machine learning model. These features were built using the initial data procured from different data streams of the company. 
\end{enumerate}
The features built during the feature engineering process represented information on a statistical level. All the private information related to employees was anonymized as per GDPR regulations. It was thoroughly ensured that features don’t track the activity of any employee on an individual level.

\subsubsection{Features scaling}
The feature scaling step followed was normalizing the values of the features using the Min-max scaler. The standard setting of the Min-max scaler was used, and the original values were scaled down a range of values between 0 to 1. During the process of making predictions on time series data, the input features from previous time steps like t-1, t-2, t-3, etc. were also fed into the supervised learning algorithm to predict the target variable.
Because the supervised learning algorithm cannot access the previous time steps automatically, the raw data set needed to be adjusted to form a supervised learning problem using the sliding window method \cite{brownlee2017introduction}.
%The distinguishing factor of the time series data is that they are observations ordered by time and 
By using the sliding window method, the order of the time series can remain maintained. 
Although that in comparison to the univariate time series, the multivariate time series has multiple observations recorded for the same time, the sliding window method can also be extended to the multivariate time series data in order to prepare it for a supervised learning algorithm. 
%Min-max scaler was used to ensure that some feature values that have a significantly higher magnitude than others do not get higher importance by the machine learning algorithm while training.
%The scaled data set was converted to suit a supervised learning problem. 
%Approaching the time series forecasting as a supervised learning problem gives the freedom to use a wide range of linear and nonlinear supervised learning algorithms. 
\begin{figure*}[!h]
\centering
\begin{minipage}{.50\textwidth}
  \centering
  \includegraphics[width=0.7\linewidth]{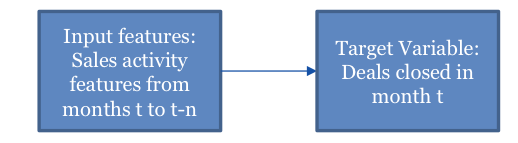}
  \captionof{figure}{Framework of the supervised learning problem for this project.}
  \label{fig:sup-problem}
\end{minipage}%
\end{figure*}

The initial data set consisted of multiple sales activities of the company (multivariate time series data) and the corresponding deals closed on a monthly level. The data set was readjusted to use the deals closed in a particular month as the target variable which can be referenced to as current time step t. The corresponding features were formed by the sales activity of the current month t, and the lagged variables of the sales activity from previous months t-n where n represents the number of months from previous time steps.  In the time series terminology, the data set has been reframed for a single step forecasting using multivariate time series with different lags. Figure \ref{fig:sup-problem} summarizes the framework of the supervised learning problem formed for this project. 
%The following two examples can be used to solidify the understanding:

%\textbf{Example 1}: When the data set is lagged by 1 month, the data instance from 31-07-2017 would have deals closed from 31-07-2017 as the target variable. The sales activity features from 31-07-2017 and 30-06-2017 would form the input feature variables. 

%\textbf{Example 2}: When the dataset is lagged by 2 months, the data instance from 31-07-2017 would have the deals closed from 31-07-2017 as the target variable. The sales activity features from 31-07-2017, 30-06-2017, and 31-05-2017 would form the input variables. 
%The same logic can be extended for identifying the target variable and input features for the data instance from 31-07-2017 as the lag of the month's increases. 
\subsubsection{Data set prepared for the time series forecasting}
The final data sets (with the lag 1-5) for time series forecasting were prepared by building the raw data set on the basis of the interview results, feature scaling, and creating lag variables using the sliding window method. 
Data streams discovered during the interview process contain ten years’ worth of data, from 2010 – 2019 related to company’s sales activities. The data streams together with the features used for the time series forecasting are presented in the Table \ref{tab:time-forecast}.
%The first column contains the name of the features, the second column contains the data sources of the company utilized to create the features and the third column contains the category of the feature variable. 
The features were accumulated on a monthly level, formed a regular time series and represented information on a statistical level.
\begin{table}[!h]
\scriptsize
\centering
\caption{Features for time series forecasting.}
\begin{tabular}{l|ll}
\hline
Feature                                           & Source                               & Category           \\ \hline
Meetings by sales employees                       & Google calendar, Employee competence & Numerical Variable \\ \hline
Meetings of sales employees with top clients      & Google calendar, Revenue data        & Numerical Variable \\ \hline
Meetings about top clients in the company         & Google calendar                      & Numerical Variable \\ \hline
Physical meeting of sales employees with   client & Credit card data                     & Numerical Variable \\ \hline
Time spent in meetings by sales employees         & Company time logger                  & Numerical Variable \\ \hline
Deals closed                                      & Company CRM (Hubspot)                & Numerical Variable \\ \hline
\end{tabular}
\label{tab:time-forecast}
\end{table}

\begin{figure*}[!h]
\centering
\begin{minipage}{.80\textwidth}
  \centering
  \includegraphics[width=0.95\linewidth]{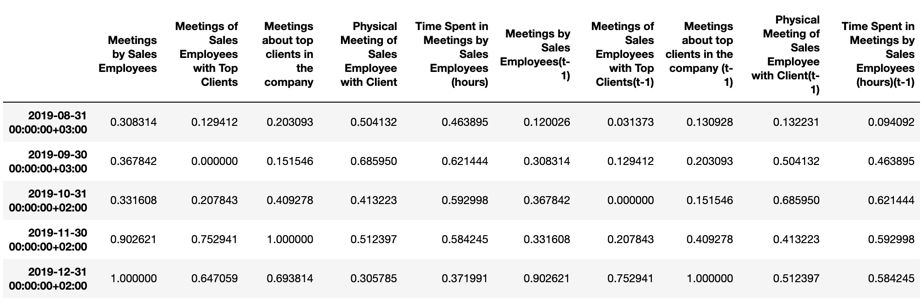}
  \captionof{figure}{A sample from the data set with a lag of 1.}
  \label{fig:dataset-prep}
\end{minipage}%
\end{figure*}

Figure \ref{fig:dataset-prep} depicts a sample of the five data instances from the data set with a lag of 1.  The final data set with a lag of 1 contains the features from time instance t (or t-0) and the features from time instance t-1 (lag of 1), represented with an additional t-1 in the column name. For example, the data instance with index “30-09-2019” contains feature values from September 2019 and the feature values from August 2019. This helps in capturing the temporal relationships between the feature values and target variable while training the machine learning model. The same logic can be extended to visualize the data sets with a lag of 2, 3, 4 and 5. 
%From the Figure \ref{fig:dataset-prep} it can also be seen that the feature values were normalized using feature scaling. 

\subsection{Time series forecasting using support vector regression}
After understanding the features, we approached the time series forecasting as a supervised learning problem using machine learning. 
%Approaching the time series forecasting as a supervised learning problem gave us the freedom to use a wide range of linear and nonlinear supervised learning algorithms. 
The reason for this approach, as mentioned by Bwonlee in \cite{brownlee2017introduction} is that machine learning algorithms can capture nonlinear relationships between the feature variables and the target variable whereas the traditional time series forecasting techniques are capable of only capturing linear relationships between the features and the target variable.  
Brownlee also highlighted \cite{brownlee2017introduction} that classical machine learning algorithms are incapable of automatically detecting the temporal dependencies in data while making predictions. %Machine learning algorithms can also be used with both univariate and multivariate time series data.
%That is why it becomes important to lag the time series manually (approaching time series forecasting as supervised learning) during forecasting or classification tasks with machine learning. 

%An exception to this is an advanced deep learning algorithm like Long short-term memory (LSTM).
%The second reason is that making a prediction on time series data is just one part of this project. The main focus lies in interpreting the model built on time series data to draw valuable insights. As found during the literature survey in [4, p. 191], various model agnostic explainability techniques are also incapable of automatically understanding the temporal dependencies of the time series data. As shown by Mokhtari et al. [21, p. 168-170], even if an advanced deep learning algorithm like LSTM capable of detecting temporal dependencies is used, model agnostic explainability methods would be unable to understand these temporal dependencies while generating explanations. That is why Mokhtari et al. [21, p. 170] created lag variables for their time series manually before feeding it to LSTM.

The data set prepared for forecasting was fed into the support vector regression algorithm. The choice of the machine learning model to be used was based on the amount of data available. 
Because the size of the data set was not big enough only the classical machine learning model algorithms were considered. 
%As noted by Chniti et al. in [13, p. 80], a deep learning algorithm like LSTM due to their state-of-the-art architecture are considered perfect for dealing with sequential data like time series.  
%Since LSTM requires large amounts of data, they were not used during this study.
Moreover, the comparative study by Chniti et al. \cite{chniti2017commerce} found that the performance of the SVR model was comparable and even better than for example a deep learning algorithm LSTM in some cases. 
We performed the time series forecasting using support vector regression (SVR) in the following way:
\begin{enumerate}
\item Training and test data set: Five different data sets containing lag variables from 1 to 5 were created using the sliding window method. The data set with lag 1 contained input features from time instance t and time instance t-1. The data set with lag 2 contained input features from time instance t, time instance t-1, and time instance t-2. The same logic can be extended to understand the structure of the data sets with lags 3, 4 or 5. After the data sets with different lags were created, each data set was split into training (80\%) and test set (20\%). 
%The common practice is to shuffle the data before creating the training and test set, however, this methodology does not work with time series data as the order of sequencing needs to be preserved while creating training and test set.
\begin{table}[!h]
\scriptsize
\centering
\caption{The set of hyperparameters passed to GridsearchCV.}
\begin{tabular}{l|l}
\hline
Kernel  & Linear, RBF            \\ \hline
C       & 0.1, 1.5, 10, 25, 50   \\ \hline
Gamma   & 1e-2, 1e-3, 1e-4, 1e-5 \\ \hline
Epsilon & 0.1, 0,2, 0.3, 0.5     \\ \hline
\end{tabular}
\label{tab:grid}
\end{table}
\item Tuning the hyperparameters of the SVR model: The hyperparameters of the SVR model were thoroughly tuned during the training process for each of the five data sets with different lag variables. The hyperparameters namely C, epsilon, gamma, and kernel were trained using the technique of GridsearchCV. 
%The default cross-validation parameter is the 3-step cross-validation, which shuffles the order of the instances during the training process. However this kind of technique does not work for time series where the instances are time-dependent.
In order to overcome the limitation of normal cross-validation, 4 step time series cross-validation in GridsearchCV was used. A list of hyperparameters presented in the table \ref{tab:grid} was passed to GridsearchCV. The best performing hyperparameters for the SVR model trained on each of the five data sets were obtained.

% \item Performance Evaluation – The performance of the SVR models built on the training data for each of the five data sets with different lag variables was evaluated on the test data set. This was done by using the mean absolute percentage error as the performance metric. Mean absolute percentage error is one of the most popular used metric to evaluate time series predictions with the advantage of being intuitive and easily understandable by humans [25]. 
\end{enumerate}

\subsection{Local explanations with SHAP and LIME}
The predictions made by the SVR model were interpreted using LIME and SHAP explainability methods to understand the influence of features on the predictions at a local level. 
The best performing model based on the least mean absolute percentage error was interpreted using the python-based implementation of Kernal SHAP \cite{shap2019} and LIME \cite{Marco2019}. Ten prediction instances from the test data set of the best performing model were chosen and the local explanation for each of them was obtained using both SHAP and LIME. 
The explanations with LIME were generated using the following steps:
\begin{enumerate}
\item The LIME explainer was prepared by passing all the values from the training data set. The parameters used while preparing the LIME explainer were mode and feature names. The mode was set to regression as the machine learning problem was of regression type. The name of the features from the training data set was passed to the parameter feature names. 
\item The LIME explainer was used to produce the explanation for the prediction instances from the test data set. Two parameters were passed to the explainer. The first parameter were the feature values of the prediction instance that needed to be explained and the second one was the trained machine learning model along with the predict attribute. 
\item A human friendly explanation was displayed using LIME explainer’s Pyplot figure function. 
\end{enumerate}
The explanations with SHAP was generated using the following steps:
\begin{enumerate}
\item The SHAP Kernel explainer was prepared by passing the trained machine learning model with the predict attribute and all the values from the training data set. 
\item The values from the test data set were passed through the prepared SHAP Kernel explainer. This helped to obtain the SHAP values for all the features in each prediction instance of the test data set. 
\item A human-friendly explanation was displayed using the SHAP’s summary plot attribute. SHAP’s summary plot attribute produced the explanation for the required prediction instance from the test data and simultaneously displayed it in form of a human-understandable plot. The first parameter passed to SHAP’s summary plot attribute to produce the explanation were the SHAP values of the prediction instance that needed to be explained (the values were obtained in step 2) and the second one were the feature values for prediction instance that needed to be explained. 
\end{enumerate}

\subsection{Evaluation of Explanations} In order to evaluate the quality of the explanations produced by LIME and SHAP a human evaluation study was conducted. The human evaluation study in \cite{malhi2020explainable} formed the basis for the human evaluation study conducted during the present study. The goal of the study was to evaluate if the explanations aid humans in understanding the decision-making process of the underlying machine learning model. We decided to choose the human grounded evaluation of explanations involving ordinary humans and a simple task. Designing an experiment for application grounded evaluation was infeasible during this project due to constraints of time and resources costly in terms of time and effort.

%\subsection{Software technologies}
%The code for this project involved the use of two programming languages, Python 3.7.6 and Structured Query Language (SQL) version 15.0. 
%SQL was used to procure the data from the company servers stored in Amazon Web Services (AWS) cloud service called S3. The interface used to interact with the S3 is called AWS Athena. Athena is a tool that provides the ability to interact and procure the data in a S3 using SQL. 
%Python was used for data cleaning, data preparation, machine learning implementation, and interpreting the machine learning model. The integrated development environment (IDE) used to write the code was Jupyter Notebook. The exact python libraries that were used during this project were: Numpy – Library for performing mathematical operations on the data; Pandas – Library for data preprocessing and data analysis; Scikit-learn – Library for machine learning; Matplotlib and Seaborn – Libraries for graphical visualizations and plots; SHAP – Library used for SHAP explainability; LIME – Library used for LIME explainability; SPSS – Software for statistical analysis; Google Docs – Human evaluation of explainability.

\section{Human Evaluation Study}

\subsection{Demographics of the study participants}

For the user-centric studies, we got the participants from the University’s environment which means that most of the participants have at least a Bachelor’s university degree. Other than that, no particular control was exercised over the study participants. The study was conducted with a total of 60 human participants (users), 20 for the LIME, 20 for the SHAP, and 20 for noXAI setting. The only thing that was ensured was that a set of 20 unique participants interacted with the LIME, SHAP, and noXAI applications, bringing the total count of the participants for the study to 60. This was done to ensure that the results for each application were unbiased as seen in \cite{hase2020evaluating}, \cite{malhi2020explainable} and \cite{nguyen2018comparing}. 

\begin{table}[!h]
\scriptsize
\centering
\caption{Demographic of the study participants.}
%\begin{longtable}{c|c|c|c|c|c|c|c|c|c|c}
\begin{tabular}{l|l|l|l|l|l|l|l|l|l|l}
\hline
Methods & 
{}{}Total & \multicolumn{3}{l|}{Gender}      & \multicolumn{3}{l|}{Highest Degree}
& \multicolumn{2}{c|}{STEM} & {\begin{tabular}[c]{@{}l@{}}Age\\ (years)\end{tabular}}   
    \\ \cline{3-10}
                         &                        & Male        & Female     & Others     & \begin{tabular}[c]{@{}l@{}}Master’s  \\ degree\end{tabular} & \begin{tabular}[c]{@{}l@{}}Bachelor’s\\ degree\end{tabular} & \begin{tabular}[c]{@{}l@{}}High\\ School\end{tabular} & Yes          & No         &                                                                                                                                                       \\ \hline
noXAI                    & \textit{20}            & 12          & 8          & 0          & \textit{13}                                                 & \textit{7}                                                  & \textit{0}                                            & \textit{12}  & \textit{8} & \textit{\begin{tabular}[c]{@{}l@{}}20 (2), 22, 24, \\ 25 (2), 26 (2), \\ 29 (1), 30 (1),\\ 31 (2), 33,\\ 34(2), 38,\\ 41, 47, 50, \\ 51\end{tabular}} \\ \hline
LIME                     & \textit{20}            & \textit{11} & \textit{8} & \textit{1} & \textit{8}                                                  & \textit{10}                                                 & \textit{2}                                            & \textit{17}  & \textit{3} & \textit{\begin{tabular}[c]{@{}l@{}}18, 20, 21, \\ 23 (3), 24 (5), \\ 25(2), 27, \\ 28 (2), 29, 31, \\ 32, 55\end{tabular}}                            \\ \hline
SHAP                     & \textit{20}            & \textit{15} & \textit{5} & \textit{0} & \textit{10}                                                 & \textit{7}                                                  & \textit{3}                                            & \textit{15}  & \textit{5} & \textit{\begin{tabular}[c]{@{}l@{}}21 (2),  22 (2), \\ 24 (2), 25 (3),\\ 26 (4), 27 (3), \\ 28, 29, 34, \\ 35\end{tabular}}                           \\ \hline
%\end{longtable}
\end{tabular}
\label{tab:demo}
\end{table}

Table \ref{tab:demo} summarizes the demographics of the study participants of LIME, SHAP, and noXAI studies. 
From the overall analysis it can be deduced that for noXAI and LIME case studies, the gender ratio of the participants was almost equal. However, for the SHAP case study, the distribution was biased more towards the male gender. It can also be seen that for the noXAI case, the majority of participants were split between the age bracket of 21-30 and 31-40 years. However, for the LIME and SHAP cases, the majority of participants were from the age group of 21-30 years. The majority of participants for all three studies held a Bachelor’s or Master’s degree and were from the STEM background. 

\subsection{Structure of the study} 
\begin{figure*}[!h]
\centering
\begin{minipage}{.55\textwidth}
  \centering
  \includegraphics[width=0.9\linewidth]{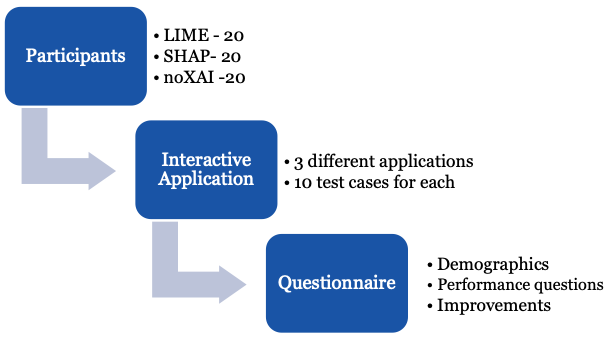}
  \captionof{figure}{Structure of the study.}
  \label{fig:study-structure}
\end{minipage}%
\end{figure*}
Three different interactive series of tests in the web-based survey were built. The first one was used to evaluate the explanation generated by LIME, the second one was used to evaluate the explanation generated by SHAP and the third one acted as a baseline where no explainable AI (noXAI) method was used. The normalized version of the input data was presented to all the 60 participants due to privacy reasons.  Each application presented every participant with 10 cases of different prediction instances from the test data set. The 10 prediction cases were shown in the same order with their respective LIME, SHAP, and noXAI explanation to avoid any kind of bias that could have been caused due to their order. The details are as follows:

\textbf{1. The LIME/SHAP based survey test setting} consisted of the normalized input features of the machine learning model, output/prediction of the model, and explanation generated by LIME/SHAP highlighting the reasoning behind the prediction made by the machine learning model. The input during this project was the sales activity features with lags on a monthly level and the output was the sales deals closed on a monthly level. Each of the 20 users was presented with 10 different cases of input, output, and the corresponding LIME/SHAP explanation. At the end of each case, the users were asked if the explanation helped them in understanding the prediction in a more understandable way. The response of the users was recorded as a binary, yes or no.

\textbf{2. The noXAI survey test setting} consisted of the input to the machine learning model, output, and no explanation (no XAI). Each of the users was presented with 10 different cases of input and output and no explanation about the decision-making process of the underlying model was given. At the end of each case, the users were asked if they were able to understand the prediction. The response of the users was recorded as a binary, yes or no.

\begin{figure*}[!h]
\centering
\begin{minipage}{.50\textwidth}
  \centering
  \includegraphics[width=0.6\linewidth]{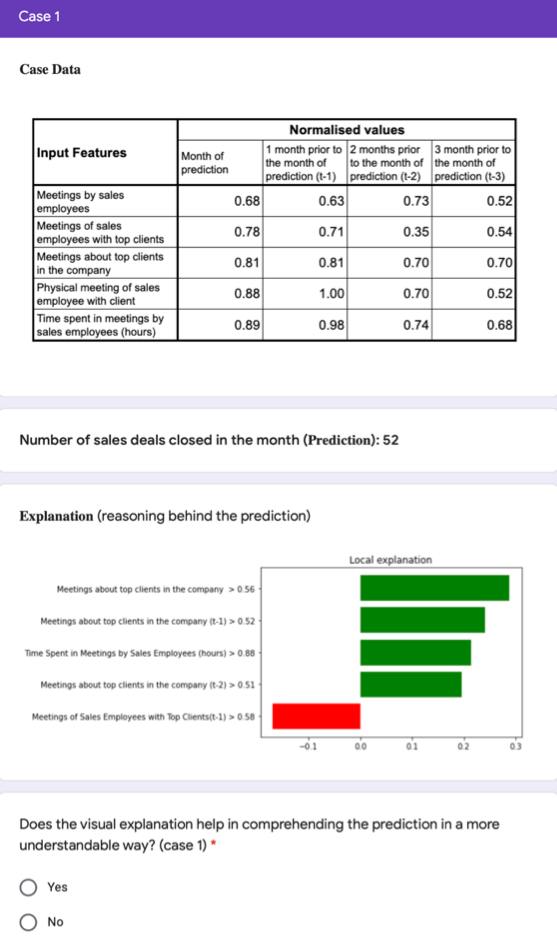}
  \captionof{figure}{One case instance from the LIME setting.}
  \label{fig:LIMEsurvey}
\end{minipage}%
\begin{minipage}{.512\textwidth}
  \centering
  \includegraphics[width=0.612\linewidth]{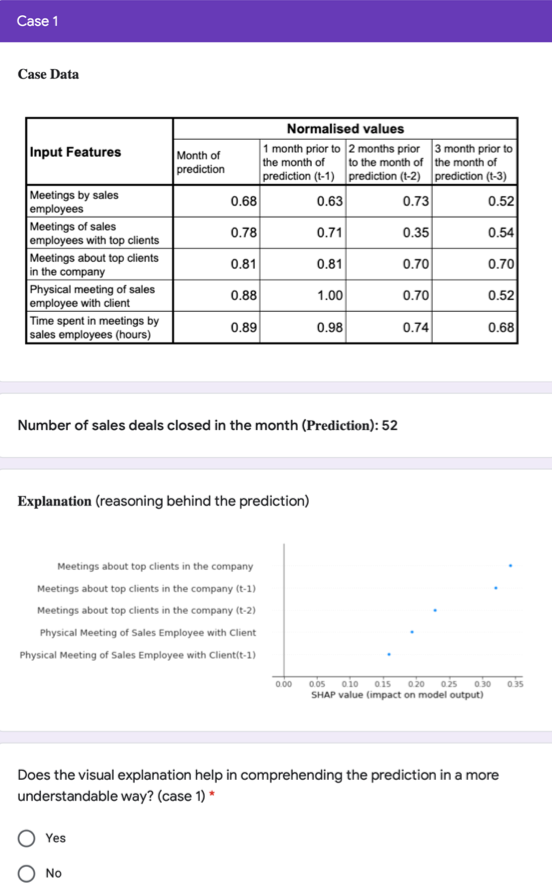}
  \captionof{figure}{One case instance from the SHAP setting.}
  \label{fig:SHAPsurvey}
\end{minipage}%
\end{figure*}

\begin{figure*}[!h]
\centering
\begin{minipage}{.50\textwidth}
  \centering
  \includegraphics[width=0.7\linewidth]{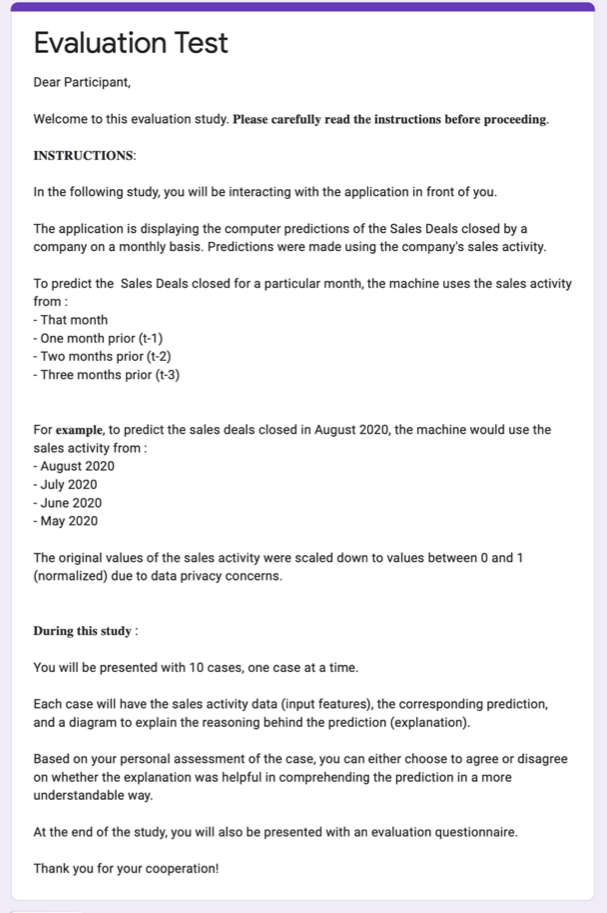}
  \captionof{figure}{Instructions in the LIME and SHAP user study.}
  \label{fig:LIMESHAPsurvey}
\end{minipage}%
\begin{minipage}{.50\textwidth}
  \centering
  \includegraphics[width=0.7\linewidth]{figures/instructions.png}
  \captionof{figure}{Instructions in the noXAI user study}
  \label{fig:NOXAIsurvey}
\end{minipage}%
\end{figure*}

Figures \ref{fig:LIMEsurvey}, \ref{fig:SHAPsurvey}, \ref{fig:NOXAIsurvey} bellow are depicting the examples taken from each of the survey test setting (LIME, SHAP, noXAI) built for the human evaluation of explanations \footnote{See Appendix for the Figures depicting the instructions presented to the participants at the beginning of the evaluation test in the LIME, SHAP and noXAI and the example from the noXAI survey test setting.}.

\subsection{Study Protocol} The study was conducted remotely, and the protocol outlined in the Figure \ref{fig:study-structure} was followed during the study:
\begin{enumerate}
\item The links of the survey test settings were shared with the different sets of users. The introductory page of the user study thoroughly discussed the time series forecasting case along with all the relevant instructions related to the tasks that had to be performed by the users.

\item Additional questions and queries regarding the cases were answered over a call. However, it was made sure that no additional information about the LIME and SHAP explanations was shared since it is important for the users to be able to understand the explanations on their own. Doing so would have defeated the purpose of the study and could have added bias in the user’s mind. 

\item After the instructions, the users went through the 10 cases in their respective LIME, SHAP, and noXAI survey test setting. At the end of each case in the LIME and SHAP setting, the users were asked if the explanation helped them in understanding the prediction in a more understandable way and the response was recorded as a yes or no. At the end of each case in the no XAI setting, the users were asked if they were able to understand the prediction made and the response was recorded as a yes or no.

\item After the users went through all the cases in the survey test setting, they were presented with a questionnaire involving the additional questions about their demographics, assessment of the explanations, and user experience with the user study test setting. 
\begin{itemize}
\item In all three survey test settings the users were asked to rate their satisfaction level with the user experience on a Likert scale of 0-5 and were encouraged to suggest possible improvements to the user experience.

\item In the LIME and SHAP survey test settings the users were asked to rate their satisfaction level with the explanations on a Likert scale of 0-5 and they had to answer if the explanations were good enough for them to trust the predictions (input: yes or no). Additionally the users were asked to suggest possible improvements to the explanations that can help improve their understanding.

\item In the noXAI setting the users were asked if they could trust the predictions without appropriate explanations (input: yes or no) and if they think that the predictions would be more satisfying/trustable if they were supported by explanations (input: yes or no). The users were also asked to suggest the kind of explanations that could be helpful.
\end{itemize}
\end{enumerate}

\subsubsection{Methods for hypotheses and correlation analyses}
\medskip
\textbf{Hypotheses.} The aim of this HCI study was to evaluate the following hypothesis:

\textbf{1. Ha}: The number of times the study participant is able to understand the prediction with LIME explanation will be greater than in the case with no explanation. (LIME $>$ noXAI)

\textbf{2. Hb}: The number of times the study participant is able to understand the prediction with SHAP explanation will be greater than in the case with no explanation. (SHAP $>$ noXAI)

\textbf{3. Hc}: The number of times the study participant is able to understand the prediction with LIME explanation will be greater than in the case with SHAP explanation. (LIME $>$ SHAP)

\begin{figure*}[!h]
\centering
\begin{minipage}{.60\textwidth}
  \centering
  \includegraphics[width=0.9\linewidth]{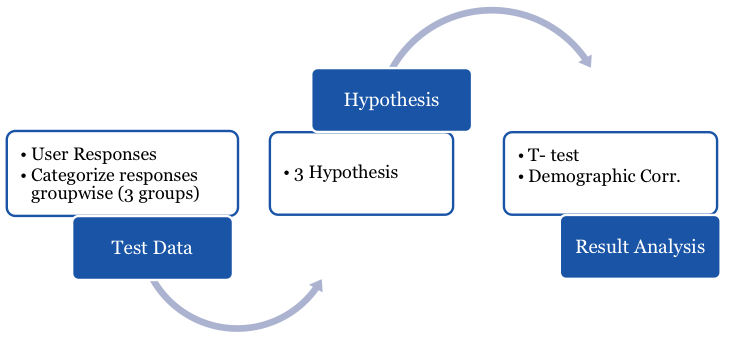}
  \captionof{figure}{Analysis method used for hypotheses evaluation .}
  \label{fig:hypo1}
\end{minipage}%
\end{figure*}

The hypotheses testing was conducted in order to test if the null hypotheses (Ha0, Hb0, Hc0), the negation of the aforementioned three hypotheses could be rejected.
In order to understand the usefulness of the explanations for the human participants the comparative tests were conducted between the three different sets of application users (SHAP, LIME, and noXAI). The hypotheses testing was conducted using a statistical two-sample t-Test. Figure \ref{fig:hypo1} summarizes the method for hypotheses analysis. 
We also measured the correlation between the demographics of the participants and their ability to understand the prediction in the LIME, SHAP, and noXAI case studies using the Spearman's rank correlation coefficient.
Spearman's correlation was chosen because it captures the monotonic relationship between the variables instead of just a linear relationship and also works well with categorical variables like gender \cite{artusi2002bravais}. 

\section{Results}
\subsection{Results of the time series forecasting}
Table \ref{tab:performance} summarizes the performance of time series forecasting models built with SVR. 
%The first column presents the amount of lag in the dataset used to train SVR, the second column presents the mean absolute percentage error on the test dataset and the third column presents the best set of hyperparameters for that particular model.
The machine learning model trained using the data set with a lag of 3 (presented in bold) was the best performing model with a mean absolute percentage error of 9.29\%. The best model was further interpreted using the model agnostic techniques, LIME and SHAP.

\begin{table}[!h]
\scriptsize
\centering
\caption{Performance summary of support vector regression models.}
\begin{tabular}{l|l|l}
\hline
Lag in the data set & Mean absolute percentage error & Best hyperparameters                            \\ \hline
1                  & 11.32 \%                       & C:1.5,   epsilon: 0.2, gamma: 0.01, kernel: RBF \\ \hline
2                  & 10.72 \%                       & C:1.5,   epsilon: 0.1, gamma: 0.1, kernel: RBF  \\ \hline
3                  & 9.29 \%                        & C:1.5, epsilon: 0.1, gamma: 0.1, kernel: RBF    \\ \hline
4                  & 11.24 \%                       & C:10,   epsilon: 0.1, gamma: 0.01, kernel: RBF  \\ \hline
5                  & 11.41 \%                       & C:1.5,   epsilon: 0.1, gamma: 0.1, kernel: RBF  \\ \hline
\end{tabular}
\label{tab:performance}
\end{table}

\subsection{Explanations generated with LIME and SHAP} 
\begin{figure*}[!h]
\centering
\begin{minipage}{.70\textwidth}
  \centering
  \includegraphics[width=0.9\linewidth]{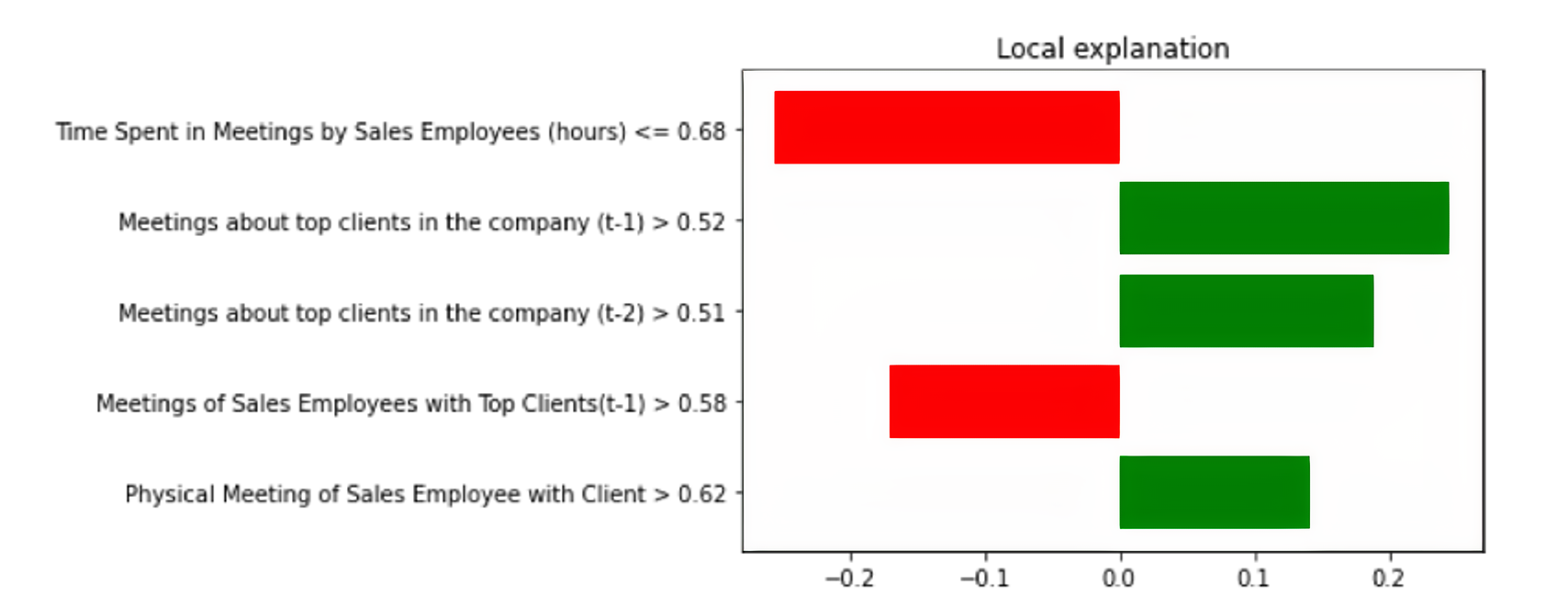}
  \captionof{figure}{Explanation generated by LIME.}
  \label{fig:LIMEexp}
\end{minipage}%
\end{figure*}

On top of the best performing SVR model LIME and SHAP explanations were generated. Figure \ref{fig:LIMEexp} depicts the LIME explanation for one prediction instance from the test data set. The prediction instance is the number of the sales deals closed in December 2018.
The x-axis of the LIME explanation (Figure \ref{fig:LIMEexp}) represents the magnitude of feature impact on the model output/prediction and the y-axis represents the name of the feature variables. The explanation only contains the top 5 features. The features are arranged (top to bottom) in the descending order of their importance (positive or negative). The red-colored bar signifies that a particular feature had a negative impact on the model output and the green colored bar signifies that a particular feature had a positive impact on the model output. The length of the bar is representative of the magnitude of impact. The LIME explanation also tries to provide a reasoning behind the positive or negative impact of a feature on the model output in the form of cut off values. They are auto generated by LIME and can be seen in the y-axis along with the names of the feature variables. 

\begin{figure*}[!h]
\centering
\begin{minipage}{.70\textwidth}
  \centering
  \includegraphics[width=0.9\linewidth]{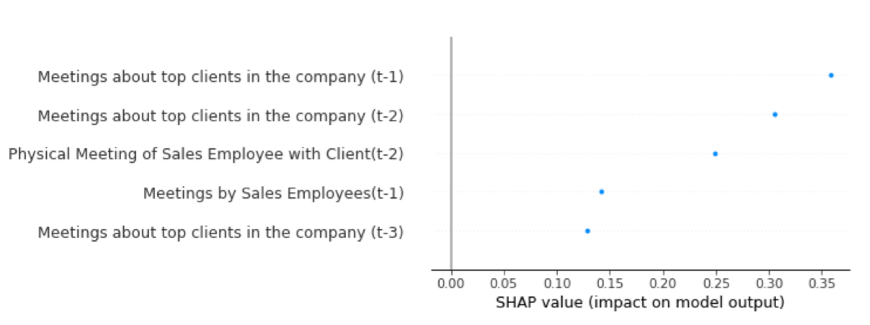}
  \captionof{figure}{Explanation generated by SHAP.}
  \label{fig:SHAPexp}
\end{minipage}%
\end{figure*}

%The most important feature contributing towards the prediction of sales deals closed in December 2018 is “Time Spent in Meetings by Sales Employees (hours)”. The feature had a very high negative impact because its value (normalized) was less than 0.68. The second most important feature is “Meetings about top clients in the company” held in November 2018 (represented with t-1). This feature had a positive impact on the predicted number of sales deals closed in December 2018 because its value (normalized) was greater than 0.52. 
%Similarly, the third most important feature is “Meetings about top clients in the company” held in October 2018 (represented with t-2). The feature had a positive impact on the predicted number of sales deals in November 2018 because its value (normalized) was greater than 0.51. 
%However, sometimes all elements of the LIME explanation may not be easily explainable, at least by lay humans. For example, the fourth most important feature according to LIME explanation is “Meetings of Sales Employees with Top Clients” held in October 2018 (represented with t-1). According to LIME, the feature had a negative impact on the predicted number of sales deals closed in December 2018 because its value (normalized) was greater than 0.58. It seems difficult to interpret the fact that a greater number of meetings with top clients led to a negative impact. An explanation of this sort would perhaps  make more sense to domain experts. 
Figure \ref{fig:SHAPexp} depicts the SHAP explanation for the same prediction instance. The x-axis of SHAP explanation also represents the magnitude of feature impact on the model output/prediction which in the case of SHAP is measured in terms of shapley values. The y-axis represents the name of feature variables. The explanation also contains only the top 5 features and the features are arranged (top to bottom) in the descending order of their importance. Unlike LIME, SHAP explanations do not show any reasoning behind the positive or negative impact of a feature on the model output. The magnitude of the impact is visually displayed with the help of dots in SHAP explanation as opposed to bars in LIME. According to the SHAP explanation, the top 5 features only had a positive impact on the prediction as opposed to the LIME explanation where 2 out of the top 5 features were shown to have a negative impact. 3 out of the top five features from SHAP and LIME explanations are the same whereas 2 are different.  

According to LIME the most important feature contributing towards the prediction of sales deals closed in December 2018 is “Time Spent in Meetings by Sales Employees (hours)”. The feature had a very high negative impact because its value (normalized) was less than 0.68. According to SHAP explanation, the most important feature is “Meetings about top clients in the company” held in November 2018.

\subsection{Quantitative analysis of the human evaluation user studies}
Table {\ref{tab:heval}} depicts the sum, mean, and median of the responses collected from the participants during the study. The results of the human evaluation study very well align with the assumptions made while designing the hypotheses Ha and Hb.  
During the LIME study, in 86 out of 200 cases, the participants reported that the LIME explanation was helpful in understanding the prediction. During the SHAP study, that was true for 82 out of 200 cases, whereas during the study where no explanation was provided to the participants, in only 7 out of 200 cases the participants said that they were able to understand the prediction. 
Similarly, we observed that in 114 out of 200 cases the participants said that the LIME explanation was not helpful in understanding the prediction which was reported in 118 out of 200 cases for the SHAP explanations. In comparison to SHAP and LIME, during the noXAI study, the participants said that they were not able to understand the prediction in 193 out of 200 cases. 

\begin{table}[!h]
\scriptsize
\centering
\caption{Results of the human evaluation user study.}
\begin{tabular}{l|l|l|l|l}
\hline
                     & Statistical Measure & LIME          & \textit{SHAP} & \textit{noXAI} \\ \hline
{}{}{Yes} & Sum                 & \textit{86}   & \textit{82}   & \textit{7}     \\ \cline{2-5} 
                     & Mean                & \textit{4.30} & \textit{4.10} & \textit{0.35}  \\ \cline{2-5} 
                     & Median              & \textit{5}    & \textit{5}    & \textit{0}     \\ \hline
{}{}{No}  & Sum                 & \textit{114}  & \textit{118}  & \textit{193}   \\ \cline{2-5} 
                     & Mean                & \textit{5.70} & \textit{5.90} & \textit{9.59}  \\ \cline{2-5} 
                     & Median              & \textit{5}    & \textit{5}    & \textit{10}    \\ \hline
\end{tabular}
\label{tab:heval}
\end{table}

Additionally from the mean and median values of the “yes” responses presented it can be seen that the LIME and SHAP explanations provided an almost equal amount of help to human participants in understanding the predictions. However, the LIME explanation does seem to be a little more helpful by a very small margin. This implies that the results support the hypothesis Hc to a very little extent. 

\subsubsection{Hypotheses analysis}

\begin{table}[!h]
\scriptsize
\begin{center}
\caption{Hypothesis analysis}
\begin{tabular}{l|l|l}
\hline
t-test & Hypothesis & \begin{tabular}[c]{@{}l@{}}p-value\\ (two-tailed)\end{tabular} \\ \hline
LIME \textgreater noXAI & Ha & \textit{0.00008***} \\ \hline
SHAP \textgreater noXAI & Hb & \textit{0.001**} \\ \hline
LIME \textgreater SHAP & Hc & \textit{0.867} \\ \hline
\end{tabular}
\begin{tablenotes}
      \small
      \item Note. *p$<$.05, **p$<$.01, ***p$<$ .001
    \end{tablenotes}
\end{center}
\label{tab:hypot}
\end{table}

Table 7 presents the p-values derived from the t-tests done for hypotheses testing. The significance level for the hypothesis testing was set at $\alpha$ = 0.05. 
The results presented in Table 7, show support for both Ha and Hb hypotheses, whereas the results do not show the significant values for supporting the hypothesis Hc. 
As presented in Table \ref{tab:heval}, the mean number of cases where LIME and SHAP explanations were helpful are almost equal. However, the mean number of cases where LIME explanation was helpful (4.30 out of 10) is slightly greater than the mean number of cases where the SHAP explanation was helpful (4.10 out of 10), which supports the hypothesis Hc to a little extent. 

\begin{table}[!h]
\scriptsize
\centering
\caption{Two sample t-Test assuming unequal variance (LIME, noXAI)}
\begin{tabular}{l|l|l}
\hline
 & LIME & noXAI \\ \hline
Mean & \textit{4.30} & \textit{0.35} \\ \hline
Standard Deviation & \textit{3.50} & \textit{1.14} \\ \hline
Variance & \textit{12.22} & \textit{0.867} \\ \hline
Observations & \textit{20} & \textit{20} \\ \hline
t Stats & \textit{4.81} & \textit{} \\ \hline
P(T\textless{}=t) two-tail & \textit{0.00008***} & \textit{} \\ \hline
\end{tabular}
\begin{tablenotes}
      \small
      \item Note. *p$<$.05, **p$<$.01, ***p$<$ .001
    \end{tablenotes}
\label{tab:t-test}
\end{table}

The results of the two-sample t-test performed in order to test the validity of the hypothesis Ha are presented in the Table \ref{tab:t-test}. The p-value and the mean value for the LIME case study indicates that the LIME explanation helped the participants in understanding the prediction in a higher extent in comparison to those having noXAI setting. The LIME explanations were helpful in 4.30 out of 10 cases on average, whereas for the noXAI case study, the participants were only able to understand the prediction in 0.35 out of 10 cases on average. 

\begin{table}[!h]
\scriptsize
\centering
\caption{Two sample t-Test assuming unequal variance (SHAP, noXAI)}
\begin{tabular}{l|ll}
\hline
                           & \multicolumn{1}{l|}{SHAP}            & noXAI          \\ \hline
Mean                       & \multicolumn{1}{l|}{\textit{4.10}}   & \textit{0.35}  \\ \hline
Standard Deviation         & \multicolumn{1}{l|}{\textit{3.972}}  & \textit{1.137} \\ \hline
Variance                   & \multicolumn{1}{l|}{\textit{15.777}} & \textit{1.292} \\ \hline
Observations               & \multicolumn{1}{l|}{\textit{20}}     & \textit{20}    \\ \hline
t Stat                     & \multicolumn{2}{l}{\textit{4.059}}                    \\ \hline
P(T\textless{}=t) two-tail & \multicolumn{2}{l}{\textit{0.001**}}                    \\ \hline
\end{tabular}
\begin{tablenotes}
      \small
      \item Note. *p$<$.05, **p$<$.01, ***p$<$ .001
    \end{tablenotes}
\label{tab:t-test2}
\end{table}

Table \ref{tab:t-test2} presents the results of the two-sample t-test performed to test the validity of hypothesis Hb. The p-value and the mean value for the SHAP case study indicates that the SHAP explanation helped the participants in understanding the predictions in a greater extent in comparison to those having noXAI setting. The SHAP explanations were helpful in 4.10 out of 10 cases on average, whereas for the noXAI case study, the participants were only able to understand the prediction in 0.35 out of 10 cases on average.

\begin{table}[!h]
\scriptsize
\centering
\caption{Two sample t-Test assuming unequal variance (LIME, SHAP)}
\begin{tabular}{l|ll}
\hline
                           & \multicolumn{1}{l|}{LIME}            & SHAP            \\ \hline
Mean                       & \multicolumn{1}{l|}{\textit{4.30}}   & \textit{4.10}   \\ \hline
Standard Deviation         & \multicolumn{1}{l|}{\textit{3.496}}  & \textit{3.972}  \\ \hline
Variance                   & \multicolumn{1}{l|}{\textit{12.221}} & \textit{15.777} \\ \hline
Observations               & \multicolumn{1}{l|}{\textit{20}}     & \textit{20}     \\ \hline
t Stat                     & \multicolumn{2}{l}{\textit{0.169}}                     \\ \hline
P(T\textless{}=t) two-tail & \multicolumn{2}{l}{\textit{0.867}}                     \\ \hline
\end{tabular}
\begin{tablenotes}
      \small
      \item Note. *p$<$.05, **p$<$.01, ***p$<$ .001
    \end{tablenotes}
\label{tab:t-test3}
\end{table}

Table \ref{tab:t-test3} presents the results of the two-sample t-test performed to test the validity of hypothesis Hc. 
We can observe that there is no statistically significant difference between users having LIME and users having SHAP explanations.

\subsubsection{Demographics' correlation analyses}
We measured the correlation between the demographics of the participant and their ability to understand the prediction in the LIME, SHAP, and noXAI case studies. The correlations analyses are presented in the Table \ref{tab:demo-corr}. There were no significant correlations between most demographic attributes of the participants and their ability to understand the prediction. 
However, some moderately high correlations were discovered between the age, gender, and STEM attributes of the participants and their ability to understand the prediction in the LIME case study. 

\begin{table}[!h]
\scriptsize
\centering
\caption{Demographic correlations}
\begin{tabular}{l|l|l|l|l|l}
\hline
 & Age & Gender & Education & STEM & \begin{tabular}[c]{@{}l@{}}Knowledge \\ about XAI\end{tabular} \\ \hline
noEXP (correlation) & \textit{0.169} & -0.342 & \textit{0.307} & \textit{0.342} & \textit{-} \\ \hline
noEXP (p-value) & \textit{0.476} & 0.140 & \textit{0.187} & \textit{0.140} & \textit{-} \\ \hline
LIME (correlation) & \textit{-0.476} & -0.497 & \textit{-0.165} & \textit{0.479} & \textit{0.172} \\ \hline
LIME (p-value) & \textit{0.034*} & \textit{0.026*} & \textit{0.488} & \textit{0.033*} & \textit{0.468} \\ \hline
SHAP (correlation) & \textit{-0.261} & \textit{0.021} & \textit{-0.290} & \textit{-0.373} & \textit{-0.090} \\ \hline
SHAP (p-value) & \textit{0.266} & \textit{0.931} & \textit{0.215} & \textit{0.105} & \textit{0.706} \\ \hline
\end{tabular}
\begin{tablenotes}
      \small
      \item Note. *p$<$.05, **p$<$.01, ***p$<$ .001
    \end{tablenotes}
\label{tab:demo-corr}
\end{table}

The  significant correlations are as follows: 
\begin{enumerate}
\item A negative correlation was discovered between age and the ability of the participants to understand the predictions using the LIME explanation. This implies that as age increases, the understanding of the prediction decreases. \item A negative correlation was discovered between gender and the ability of the participants to understand the predictions using LIME explanation. This implies that a higher understanding of the predictions correlates with gender male. 
\item A positive correlation was discovered between STEM background and the ability of the participants to understand the predictions using LIME explanation. This implies that a higher understanding of the predictions correlates with participants having a STEM background. 
\end{enumerate}
The correlation between age, gender, and STEM background of the participants and their ability to understand the predictions using LIME explanation is statistically significant (p $<$ 0.05, Table \ref{tab:demo-corr}). 
It does seem quite logical that younger participants were able to make more sense out of the predictions using LIME explanation, possibly due to their affinity with computer algorithms and applications. It also makes sense that the LIME explanation was more helpful to participants with STEM background in understanding the predictions. This could be because people with STEM background are more used to graphs and numbers. The ability to understand the predictions using LIME correlates with male gender. This is possible because of the fact that there are more males in STEM-related fields.

\subsection{Qualitative analysis of of the human evaluation user studies}
Table \ref{tab:rate} presents the mean, median, and standard deviation of the subjective ratings (satisfaction) given by the participants for LIME and SHAP explanations on a Likert scale of 0 to 5. It can be observed that the mean satisfaction level with LIME explanations was higher than the mean satisfaction level of the SHAP explanations. This very well aligns with the results of the human evaluation study and the assumptions made while designing Hc. The participants found LIME explanations more helpful than SHAP explanations. 
%Figure \ref{fig:ratingLIME} depicts the distribution of the satisfaction ratings given by the participants for the LIME explanations and Figure \ref{fig:ratingSHAP} depicts the distribution of the satisfaction ratings given by the participants for the SHAP explanations.

\begin{table}[!h]
\scriptsize
\centering
\caption{Subjective ratings of the explanations.}
\begin{tabular}{l|l|l}
\hline
Statistical Measure & LIME          & SHAP          \\ \hline
Mean                & \textit{2.95} & \textit{2.6}  \\ \hline
Median              & \textit{3}    & \textit{2.5}  \\ \hline
Standard deviation  & \textit{1.70} & \textit{1.46} \\ \hline
\end{tabular}
\label{tab:rate}
\end{table}

\begin{figure*}[!h]
\centering
\begin{minipage}{.60\textwidth}
  \centering
  \includegraphics[width=0.8\linewidth]{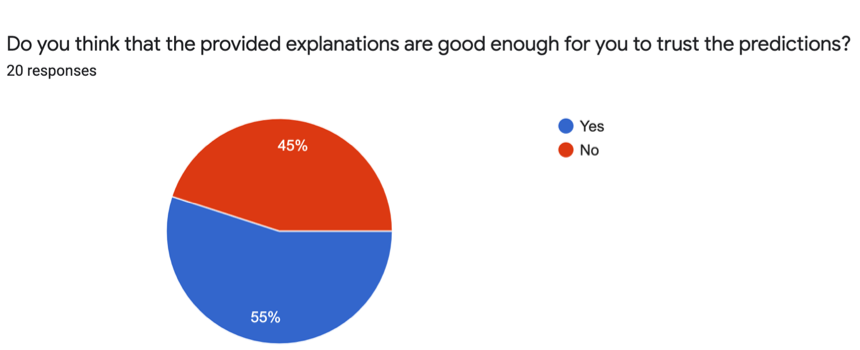}
  \captionof{figure}{Trust analysis for LIME explanations.}
\label{fig:trustLIME}
\end{minipage}%
\end{figure*}

\begin{figure*}[!h]
\centering
\begin{minipage}{.60\textwidth}
  \centering
  \includegraphics[width=0.8\linewidth]{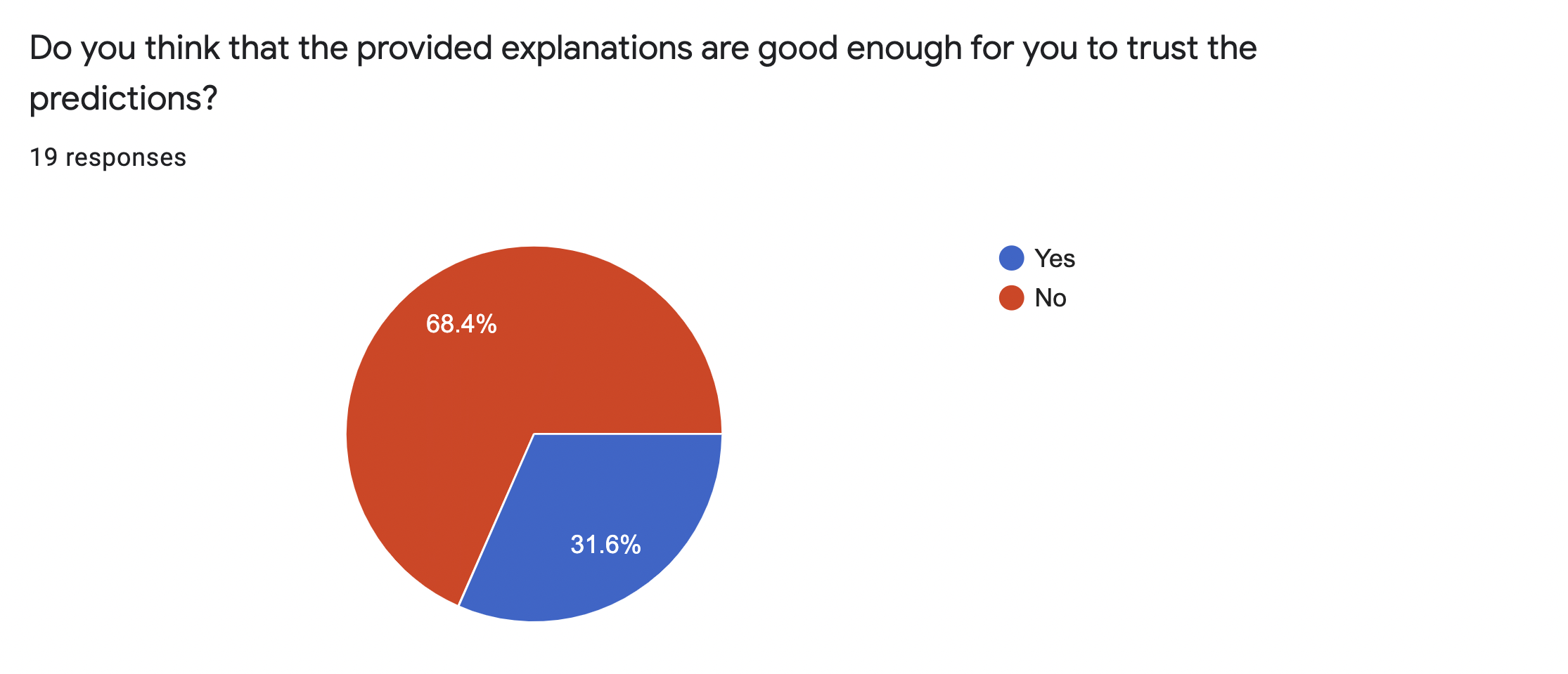}
  \captionof{figure}{Trust analysis for SHAP explanations.}
\label{fig:trustSHAP}
\end{minipage}%
\end{figure*}

\begin{figure*}[!h]
\centering
\begin{minipage}{.60\textwidth}
  \centering
  \includegraphics[width=0.8\linewidth]{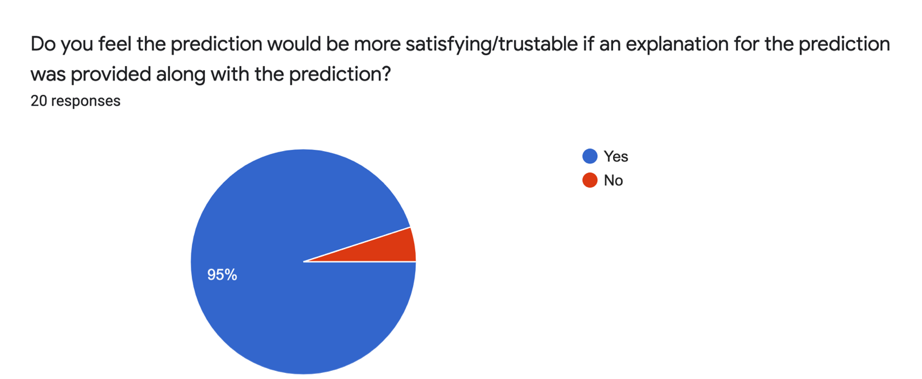}
  \captionof{figure}{Trust analysis in noXAI study.}
\label{fig:trustnoXAI}
\end{minipage}%
\end{figure*}

Further the participants of the LIME and SHAP user studies were asked if the explanations were good enough for them to trust the predictions. Figure \ref{fig:trustLIME} and \ref{fig:trustSHAP} depict the trust analysis for the LIME and SHAP explanations. As seen from the Figure \ref{fig:trustLIME}, 55\% of participants felt that the LIME explanations were good enough for them to trust the predictions. 
As seen from the Figure \ref{fig:trustSHAP}, 31.6\% of participants felt that the SHAP explanations were good enough for them to trust the predictions. 
The results of the trust analysis are consistent with the subjective rating of explanations presented in Table \ref{tab:rate} and the results of the human evaluation of explainability presented in table Table \ref{tab:heval}. LIME explanations helped the participants in understanding the predictions for a higher number of cases on average and also showed a higher average satisfaction rating as compared to SHAP. The participants of the no XAI user study were asked if the predictions would be more trustable if explanations for the predictions were provided along with them. As depicted in Figure \ref{fig:trustnoXAI}, 95\% of participants answered with a “yes”. This clearly validates the need for having explanations for the predictions made by machine learning models, be it using SHAP, LIME, or any other technique. 
%Humans have a tendency to trust something more if they understand is better.

The study participants were also asked to rate the user experience of the studies on a Likert scale of 0 to 5.  The mean, median, and standard deviation (statistical summary) of the user experience ratings are presented in Table \ref{tab:exp-rating}. From the table it can be seen that LIME received an average user experience rating of 3.60 which is greater than the average user experience rating of SHAP. These results support the results of the human evaluation study (see Table \ref{tab:heval}) and the results of the subjective ratings (see Table \ref{tab:rate}) which are also inclined towards LIME in comparison to SHAP. The difference in the user experience of the studies was heavily influenced by the LIME, SHAP, and no explanation settings because every other aspect of the studies was kept exactly the same. 

\begin{table}[!h]
\scriptsize
\centering
\caption{User experience ratings of the studies.}
\begin{tabular}{l|l|l|l}
\hline
Statistical Measure & LIME          & SHAP          & noXAI         \\ \hline
Mean                & \textit{3.60} & \textit{3.05} & \textit{3.95} \\ \hline
Median              & \textit{4.0}  & \textit{3.50} & \textit{4.0}  \\ \hline
Standard deviation  & \textit{1.53} & \textit{1.50} & \textit{0.80} \\ \hline
\end{tabular}
\label{tab:exp-rating}
\end{table}

The noXAI study received the highest average user experience rating of 3.95. This could possibly be attributed to the fact that the noXAI study did not have any kind of explanation for the predictions which in turn reduced the complexity of the study and made the user experience better as compared to SHAP and LIME user studies. 

%\begin{figure*}[!h]
%\centering
%\begin{minipage}{.60\textwidth}
%  \centering
%  \includegraphics[width=0.8\linewidth]{figures/exp-ratingLIME.png}
%  \captionof{figure}{User experience ratings for LIME study.}
%\label{fig:trustnoXAI}
%\end{minipage}%
%\end{figure*}

%\begin{figure*}[!h]
%\centering
%\begin{minipage}{.60\textwidth}
%  \centering
%  \includegraphics[width=0.8\linewidth]{figures/exp-ratingSHAP.png}
%  \captionof{figure}{User experience ratings for SHAP study.}
%\label{fig:trustnoXAI}
%\end{minipage}%
%\end{figure*}

%\begin{figure*}[!h]
%\centering
%\begin{minipage}{.60\textwi%dth}
%  \centering
%  \includegraphics[width=0.8\linewidth]{figures/exp-ratingnoXAI.png}
%  \captionof{figure}{User experience ratings for noXAI study.}
%\label{fig:trustnoXAI}
%\end{minipage}%
%end{figure*}

\section{Discussion}
In the present work we focused on explaining a machine learning-based time series forecasting model and together with that on evaluating the explanations produced with the model agnostic explainability techniques, LIME and SHAP. The machine learning model trained using the data set with a lag of 3 was based on the mean absolute percentage error the best performing model.
The selected model was further explained using LIME and SHAP. Both of the model agnostic explainability tecnhiques were able to successfully capture the temporal dependencies of underlying time series forecasting model after applying the lagging window method. The resulting explanations clearly presented the important features responsible for the prediction along with the time instance (temporal dependency).

The evaluation of the explanations produced using LIME and SHAP was conducted using the human-grounded evaluation method. 
The results of the human evaluation studies clearly proved that the explanations produced by LIME and SHAP greatly helped humans in understanding the predictions made by the machine learning model which aligned with the two of our hypotheses (Ha and Hb). The trust analysis also proved that having explanations along with the prediction can massively increase the trust of humans in the predictions made by a machine learning model. The human evaluation study results also suggested that LIME and SHAP explanations were almost equally understandable with LIME performing better but with a very small margin. This result supported the hypothesis (Hc) formulated during this project work to a little extent. Increasing the sample size for the test could help reach more statistically significant results regarding hypothesis Hc.

Further the results of demographic correlation analysis show some interesting correlations between the participant’s ability to understand the predictions using LIME explanations and their age, gender, and STEM background. Lower understanding of the predictions correlates with higher age. A higher understanding of the prediction correlates with the gender male. A higher understanding of the predictions correlates with the participants having a background in STEM.  
The trust analysis showed results in favour of LIME with a higher percentage of people saying that the explanations helped them trust the prediction of the machine learning model and the subjective satisfaction ratings by the participants were also higher for LIME.

Most of the participants were satisfied with the user experience of the LIME and SHAP user study. The user experience of the LIME study was rated higher than the SHAP study. All the comparative results between LIME and SHAP are consistent with each other and favored LIME explanations.
A few participants found the normalized feature values difficult to comprehend and perhaps the use of unnormalized values would improve the intuitiveness.
Regarding the possible improvements of LIME explanations participants stated that the explanations should be made more informative. The improvement suggestions included defining the x and y-axis in a better way, adding a legend, defining the algorithm behind the explanations, and adding informative examples. 
Participants also stated that incorporating global explanations could be helpful as well. 
Users suggested that SHAP explanations could be improved by adding the label to the x-axis and making the label of the y-axis bigger.  A major feedback regarding the explanations was about using bars or lines instead of dots to represent the SHAP values. A few participants also recommended color-coding the negative and positive SHAP values in the explanation in order to make it visually appealing and for increasing its understandability.
Participants from the study without any explanation given  were also satisfied with the given user experience and suggested that the potential explanations should be in a visual form.

\subsection {Limitations} 
The present study has a set of limitations, the most important of which are listed below:
\begin{enumerate}
\item The scope of the explainability was limited to local explainability and global explainability was not explored during this project work. Achieving global explainability using model agnostic methods is extremely difficult because it is hard for a surrogate model to mimic the full decision boundary of a complex machine learning model. However techniques like SHAP do claim that they can produce high-quality global explanations. 
\item Only two model agnostic explainability techniques, SHAP and LIME were used during this project. The scope can be extended to other sophisticated feature attribution-based explainability techniques like CIU [31] and ELI5 [32]. 
\item The complexity of the explanations generated by LIME and SHAP was limited to only the top five most important features. This could be increased to understand the effect of increased complexity on the human understanding of predictions.
\item Due to constraints of time and resources, the scope of evaluation was limited to the human evaluation of explainability which involved lay humans doing simple tasks.  The work can be extended to conduct application-grounded evaluation which involves domain experts performing real tasks. 
\end{enumerate}
 
\subsection{Future work}
To overcome the limitations addressed in the previous sub-section, the following research directions may be considered in the future:
\begin{enumerate}
\item The work done in this project can be extended to achieve explainability using other sophisticated model agnostic explainability tools like ELI5 and CIU. It would be very interesting to compare the evaluation results of ELI5 and CIU with LIME and SHAP. 
\item Furthermore in order to get a better evaluation of the explanations, they would have to be evaluated using application-grounded evaluation which involves domain experts performing tasks specific to the use of the explanations.
\item Various control elements of the evaluation studies can be changed for further analysis. Recording of time spent for completing the study could act as a very good proxy for the effort made by the participant to understand the prediction and the underlying explanation. The correlations generated using this could act as a very good evaluation metric and could lead to some interesting insights. 
\item Another suggestion is changing the complexity of the explanations. It would be interesting to see if increasing the complexity of the explanations like changing the number of features from top 5 to top 10 or 15, affects the ability of the participants to understand the predictions.
\item Due to paucity of time, the participants for the human evaluation of explainability study were limited to 60 (20 for each case). It would be logical to validate the results with a larger sample size. Moreover, increasing the sample size could help reach statistically significant results regarding hypothesis Hc (LIME $>$ SHAP). 
\item The type of human evaluation task performed in this project work was a verification task. There are other types of human evaluation tasks like forward prediction and counterfactual prediction which could potenially be explored with respect to time series. 
\item The framework used for achieving and evaluating explainability in this project work focused on a regression setting. However, this work can easily be extended to a classification setting. It would be interesting to compare the human evaluation results of explainability for a classification setting to those for a regression setting.
\end{enumerate}

\section{Conclusion}

In a world where every organization is trying to become data driven, this research presents the potential of the XAI methods that can contribute by demystifying the underlying black-box machine learning models. The focus of the present study is on highlighting how XAI techniques can be applied to a time series forecasting problem and on portraying how the quality of the explanations generated for a time series forecasting model can be evaluated with the help of the human evaluation of explainability.
The study concluded that the Local Interpretable Model-Agnostic Explanations (LIME) and Shapley Additive Explanations (SHAP) greatly aided lay humans in understanding the decision-making process of the underlying time series forecasting model and drastically improved trust in the model. From the quantitative and qualitative comparison of LIME and SHAP explanations, it appears that the human evaluators seemed to favour LIME explanations over SHAP. The difference was however negligible but perhaps an increased sample size could help solidify the comparison further.

Since the explanations were evaluated using human evaluation which involved end-users, the explanations could in the future additionally be evaluated using application-grounded evaluation which involves domain experts, performing tasks specific to the use case of the explanations. Furthermore, the scope of explainability can be extended to other XAI algorithms such as CIU and ELI5.

%Time series forecasting using machine learning has become extremely popular with applications in economics, finance, science and technology. 

%can easily be extended to any time series forecasting or classification scenario for achieving and evaluating explainability. Moreover, this work also forms a good framework for achieving and evaluating explainability in any machine learning-based regression or classification problem (supervised learning).

\section*{Acknowledgements}
The researchers would like to express their sincere gratitude towards Futurice Oy for providing the data used in this research work. Special thanks to Tuomas Syrjänen, Chief AI officer at Futurice for posing this interesting research problem and providing his valuable feedback along the way.

The research  leading  to  this  publication is partially supported by the Wallenberg AI, Autonomous Systems and Software Program (WASP), funded by the Knut and Alice Wallenberg Foundation.
%The  research  leading  to  this  publication  is  supported  by  Helsinki  Institute  for  Information  Technology  (grant9160045), under the Finnish Center for Artificial Intelligence (FCAI) unit.

 \bibliographystyle{splncs04}
 \bibliography{ref}

\section*{Appendix A:}

\begin{figure*}[!h]
\centering
\begin{minipage}{.50\textwidth}
  \centering
  \includegraphics[width=0.7\linewidth]{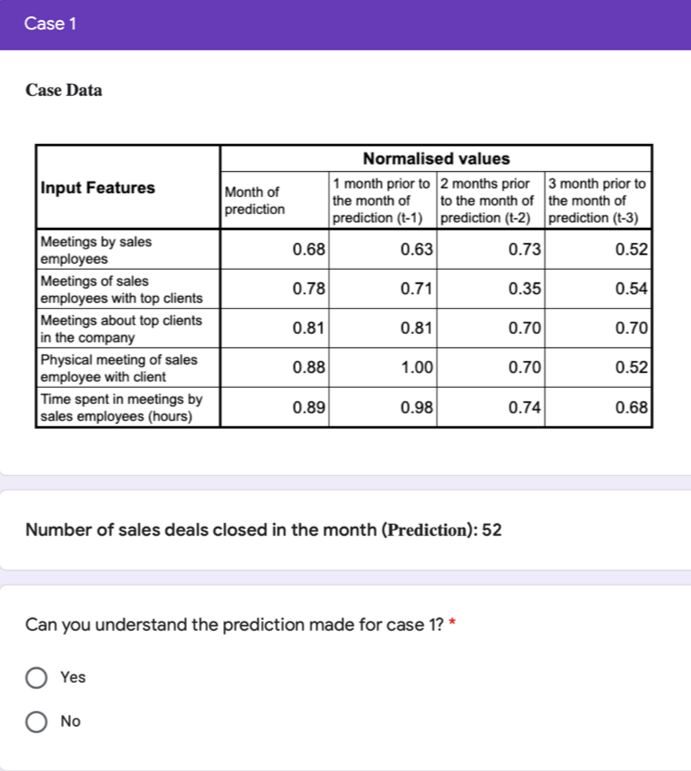}
  \captionof{figure}{One case instance from the noXAI setting.}
  \label{fig:NOXAIset}
\end{minipage}%
\end{figure*}

%\begin{figure*}[!h]
%\centering
%\begin{minipage}{.50\textwidth}
%  \centering
%  \includegraphics[width=0.7\linewidth]{figures/instructions.png}
%  \captionof{figure}{Instructions presented to the users in LIME and SHAP setting.}
%  \label{fig:expstudy}
%\end{minipage}%
%\end{figure*}

%\begin{figure*}[!h]
%\centering
%\begin{minipage}{.50\textwidth}
%  \centering
%  \includegraphics[width=0.7\line%width]{figures/noXAIstudy.png}
%  \captionof{figure}{Instructions %presented to the human %participants in the noXAI %application.}
%\label{fig:noXAIstudy}
%\end{minipage}%
%\end{figure*}

\end{document}